\documentclass{article}

\usepackage[preprint]{neurips_2026}
\usepackage{neurips_2026}

\usepackage[utf8]{inputenc}
\usepackage[T1]{fontenc}
\usepackage{hyperref}
\usepackage{url}
\usepackage{booktabs}
\usepackage{amsfonts}
\usepackage{amsmath}
\usepackage{amssymb}
\usepackage{amsthm}
\usepackage{bm}
\usepackage{graphicx}
\usepackage{multirow}
\usepackage{microtype}
\usepackage[table]{xcolor}
\usepackage{nicefrac}
\usepackage{adjustbox}
\usepackage{wrapfig}
\usepackage{makecell}
\usepackage{algorithm}
\usepackage{algpseudocode}
\usepackage{tcolorbox}
\tcbuselibrary{listings}
\usepackage{listings}

\lstdefinestyle{missbgmcode}{
  language=Python,
  basicstyle=\ttfamily\footnotesize,
  keywordstyle=\color{blue!70!black}\bfseries,
  commentstyle=\color{green!40!black},
  stringstyle=\color{orange!70!black},
  morekeywords={
    import, from, as, return, def, class,
    if, else, elif, for, while, in, not,
    and, or, True, False, None
  },
  emph={simulate_mnar_oracle_data, MissBGM, predict, fit},
  emphstyle=\color{teal!70!black},
  showstringspaces=false,
  breaklines=true,
  columns=fullflexible
}

\newtcblisting{codeblock}[2][]{
  colback=gray!3,
  colframe=gray!30,
  listing only,
  title=\textbf{#2},
  fonttitle=\small,
  coltitle=black,
  boxrule=0.5pt,
  arc=3pt,
  outer arc=3pt,
  left=6pt,
  right=6pt,
  top=6pt,
  bottom=6pt,
  listing options={
    style=missbgmcode
  },
  #1
}

\title{Missingness-aware Data Imputation via AI-powered Bayesian Generative Modeling}

\author{%
  Qiao Liu \\
  Department of Biostatistics \\
  Yale University \\
  \texttt{qiao.liu@yale.edu}%
}

\newcommand{\obs}{\mathrm{obs}}
\newcommand{\mis}{\mathrm{mis}}

\theoremstyle{plain}

\theoremstyle{definition}

\begin{document}

\maketitle

\begin{abstract}
    Missing data imputation remains a fundamental challenge in modern data science, especially when uncertainty quantification is essential. In this work, we propose MissBGM, an AI-powered missing data imputation method via Bayesian generative modeling that bridges the expressive flexibility of neural networks with the statistical rigor of Bayesian inference. Unlike existing methods that often focus on point estimates or treat the missingness mechanism implicitly, MissBGM explicitly and jointly models the data-generating and missingness mechanisms, providing principled posterior uncertainty over imputations rather than a single point estimate. We develop a stochastic optimization framework with alternating updates among missing values, model parameters, and latent variables until convergence. Our theoretical analysis shows that estimates of missing values from MissBGM converge consistently under mild assumptions. Empirically, we demonstrate that MissBGM achieves superior performance over traditional imputers and recent neural network-based methods across extensive experimental settings. These results establish MissBGM as a principled and scalable solution for modern missing data imputation. The code for MissBGM is open sourced at \url{https://github.com/liuq-lab/MissBGM}.
\end{abstract}

\section{Introduction}
Missing data is pervasive in modern machine learning and data science, arising in domains such as healthcare, social science, and industrial analytics~\citep{rubin1976inference,little2019statistical,enders2022applied,schafer2002missing,allison2009missing}. It can substantially degrade downstream analyses and lead to misleading scientific conclusions if not handled properly. This challenge is particularly acute when the missingness pattern is complex, structured, or driven by the unobserved values themselves~\citep{mitra2023learning}. Developing methods that can flexibly capture these complex dependencies while remaining statistically principled and computationally scalable is therefore a central problem for modern missing data analysis.

Classic statistical methods, such as likelihood-based models~\citep{schafer1997analysis}, expectation-maximization procedures~\citep{dempster1977maximum}, and multiple imputation frameworks~\citep{rubin1987multiple,van2011mice,carpenter2023multiple}, offer a principled way of handling missing data. However, they rely heavily on restrictive parametric models or repeated conditional modeling, which limit the scalability in modern complex and non-linear high-dimensional data. By contrast, modern deep learning methods have greatly improved imputation accuracy by leveraging the flexibility of neural networks, especially the modern generative AI architectures~\citep{yoon2018gain,mattei2019miwae,ipsen2020not,tashiro2021csdi,jolicoeur2024generating,zheng2022diffusion,zhang2025sampling}. However, those modern AI imputers primarily target point estimation and often rely on training-time masking distributions. They handle missingness mechanism implicitly or heuristically. Their multiple imputations are not automatically a principled posterior for missing values. A principled way for uncertainty quantification is essential for reliable and trustworthy data analysis~\citep{begoli2019need,abdar2021review,hossain2025beyond}.

To address these limitations, we propose MissBGM, a \underline{\textbf{Miss}}ing data imputation framework based on \underline{\textbf{B}}ayesian \underline{\textbf{G}}enerative \underline{\textbf{M}}odeling that integrates the expressive power of deep neural nets with the statistical rigor of Bayesian inference. The central idea of MissBGM is to jointly model the data-generating process, and missingness mechanism within a unified framework, providing both point estimate and uncertainty quantification for missing values. By explicitly modeling the masking process, MissBGM accounts for complex and potentially non-ignorable missingness patterns that are common in real-world data. MissBGM is powered by the flexible generative architecture to capture nonlinear dependencies in high-dimensional settings while adhering to Bayesian principles. To make this framework practical at scale, we develop an alternating stochastic optimization algorithm that iteratively updates missing values, latent variables, and model parameters in a mini-batch manner until convergence. In this way, MissBGM closes the gap between the flexibility of modern AI models and the statistical rigor of Bayesian inference, offering a powerful and scalable solution for modern missing data imputation. We summarize the contributions of MissBGM as follows.

\begin{itemize}
    \item MissBGM bridges the gap between powerful deep learning models and statistically principled Bayesian inference for missing data imputation with uncertainty quantification by explicitly modeling potentially non-ignorable missingness mechanisms.
    \item To enable MissBGM to learn the data-generating process and complex, potentially non-ignorable missingness patterns, we design an alternating stochastic optimization algorithm for efficient training and per-sample posterior inference, making the framework scalable for large-scale applications.
    \item Our theoretical analysis shows that MissBGM is consistent for the pseudo-true tempered imputation target. Moreover, we demonstrate the superior performance of MissBGM on both synthetic and real-world benchmarks with varying sample sizes and dimensions.
\end{itemize}

\section{Related work}\label{sec:related-work}

\subsection{Classical statistical imputation methods}

The classical statistical imputation methods, including likelihood-based methods~\citep{schafer1997analysis}, EM-type algorithms~\citep{dempster1977maximum}, and multiple imputation methods~\citep{rubin1987multiple,van2011mice,carpenter2023multiple}, are grounded in explicit assumptions about the data-generating process and missingness mechanisms. These methods yield coherent inference when the assumptions are properly specified. Iterative chained-equation methods are especially influential in practice, where the conditional distribution of each missing feature is iteratively estimated from the other features. One example is miceforest~\citep{wilson_miceforest_2023,doove2014recursive} which incorporates tree-based multiple imputation and often yields strong empirical performance on tabular data with nonlinear feature dependencies. Although these methods are relatively interpretable and can be competitive in practice, they generally rely on repeated conditional modeling heuristics rather than a fully generative modeling framework. More importantly, most classical statistical imputation approaches are well justified under simplified missingness assumptions and become less reliable when the masking mechanism is complex, structured, or is potentially non-ignorable~\citep{rubin1976inference}. They may also become computationally burdensome in large-scale or high-dimensional settings.

\subsection{Deep learning-based imputation methods}

%MIWAE~\citep{mattei2019miwae}, VAEAC~\citep{ivanov2018variational}, HI-VAE~\citep{nazabal2020handling}, and GP-VAE~\citep{fortuin2020gp}
%such as GAIN~\citep{yoon2018gain} and MisGAN~\citep{li2019misgan}
%CSDI~\citep{tashiro2021csdi}, TabCSDI~\citep{zheng2022diffusion}

Recent advances in deep learning, especially the emergence of generative AI models, have led to the development of imputation methods that improve imputation accuracy by learning flexible conditional structure from incomplete data~\citep{yoon2018gain,mattei2019miwae,ipsen2020not,nazabal2020handling,fortuin2020gp,tashiro2021csdi,jolicoeur2024generating,zheng2022diffusion,zhang2025sampling}. These include variational autoencoder (VAE)-based approaches~\citep{mattei2019miwae,ivanov2018variational,nazabal2020handling,fortuin2020gp,peis2022missing}, generative adversarial network (GAN)-based methods~\citep{yoon2018gain,li2019misgan}, and diffusion-style methods~\citep{tashiro2021csdi,zheng2022diffusion,jolicoeur2024generating}. For example, GAIN~\citep{yoon2018gain} formulates imputation as an adversarial learning problem, using a generator--discriminator architecture to produce plausible completions of incomplete data. ForestDiffusion~\citep{jolicoeur2024generating} combines diffusion-style generative modeling with tree-based learning, aiming to better adapt diffusion ideas to tabular data. TabCSDI~\citep{zheng2022diffusion} builds on conditional score-based diffusion models and performs imputation by iteratively denoising incomplete samples conditioned on observed entries.

Although these methods can represent rich and complex conditional distributions and can generate multiple imputations, their uncertainty quantification is generally not statistically principled under non-ignorable missingness unless the observation process is modeled explicitly~\citep{hossain2025beyond}. In particular, many methods still emphasize accurate reconstruction and either assume ignorable missingness or treat the mask simply as conditioning information or an auxiliary input, rather than as part of an explicit probabilistic missingness model. As a result, these methods provide limited interpretability regarding why entries are missing, and they do not directly support coherent inference under a specified missingness mechanism. This limitation becomes especially important in real-world settings where missingness may be structured, complex, or potentially non-ignorable.

A small but highly relevant line of work~\citep{collier2020vaes,ipsen2020not,ma2021identifiable} models the missingness process itself, and therefore is conceptually closer to MissBGM. For example, not-MIWAE~\citep{ipsen2020not} extends MIWAE~\citep{mattei2019miwae} to non-ignorable missingness settings by introducing a learned model for the mask conditional on the data, and related work~\citep{collier2020vaes} models missingness through an explicit corruption process so that latent-variable inference has principled access to mask indicators. These methods move beyond implicit mask conditioning toward explicit modeling of the observation process. However, they are typically based on variational frameworks, where posterior inference is from encoder-based approximations. In practice, such approximations can be restrictive and may yield under-estimated uncertainty.

\paragraph{Positioning of MissBGM}
MissBGM is designed at a complementary position for being both a scalable deep latent imputer and a statistically principled Bayesian model for missing data imputation. Unlike reconstruction-oriented deep imputers that condition on the mask implicitly, MissBGM explicitly models the data-generating process together with the missingness mechanism, which provides interpretation for missingness mechanisms. Unlike variational-based methods that use the same learned inference machinery for all test samples, MissBGM performs sample-specific posterior inference for latent variables and missing entries under the fitted global model. This makes inference less constrained by the chosen inference family, allows the posterior approximation to adapt to the particular observed values and mask pattern of each incomplete sample, and provides a more principled way to posterior-sampling-based uncertainty quantification.

\section{Problem setup}

We consider a dataset with $n$ samples and $p$ features, with complete data matrix $\mathbf{X} \in \mathbb{R}^{n\times p}$ and missingness indicator matrix $\mathbf{R} \in \{0,1\}^{n\times p}$.  An entry $r_{ij}=1$ indicates that $j$-th feature of $i$-th sample is observed while $r_{ij}=0$ indicates a missing entry. We partition $\mathbf{x}_i \in \mathbb{R}^p$ into $\mathbf{x}_i = (\mathbf{x}_{i,\obs}, \mathbf{x}_{i,\mis})$, where $\mathbf{x}_{i,\obs}$ and $\mathbf{x}_{i,\mis}$ are the observed and missing components of $\mathbf{x}_i$, respectively. 

Our objective is to infer the posterior distribution of the missing values given the observed data and missingness indicator matrix, which is given by
\[
    p(\mathbf{X}_{\mis} \mid \mathbf{X}_{\obs}, \mathbf{R})\,. 
\]

The missingness mechanism is typically classified into three categories~\citep{rubin1976inference,little2019statistical}: 

- \textbf{Missing Completely At Random (MCAR).} The missingness is independent of the data, so $p(\mathbf{R} \mid \mathbf{X}) = p(\mathbf{R})$. In this case, the mask $\mathbf{R}$ carries no information about the values of $\mathbf{X}$, and one can in principle ignore $\mathbf{R}$ when inferring $\mathbf{X}_{\mis}$. 

- \textbf{Missing At Random (MAR).} The missingness depends on the observed values but not on the missing ones, i.e.\ $p(\mathbf{R}\mid \mathbf{X}) = p(\mathbf{R} \mid \mathbf{X}_{\obs})$. Under MAR (and MCAR), the mask $\mathbf{R}$ is conditionally independent of $\mathbf{X}_{\mis}$ given $\mathbf{X}_{\obs}$, and hence 
\[
p(\mathbf{X}_{\mis}\mid \mathbf{X}_{\obs}, \mathbf{R}) = p(\mathbf{X}_{\mis}\mid \mathbf{X}_{\obs})\,.
\]
In other words, under MAR or MCAR (also known as {\it ignorable} missingness), valid imputation can be performed by conditioning on the observed data only. 

- \textbf{Missing Not At Random (MNAR).} The missingness mechanism depends on both observed and missing values. In the MNAR regime (sometimes called {\it non-ignorable} missingness), the mask $\mathbf{R}$ carries additional information about the missing values, and we typically have $p(\mathbf{X}_{\mis}\mid \mathbf{X}_{\obs}, \mathbf{R}) \neq p(\mathbf{X}_{\mis}\mid \mathbf{X}_{\obs})$. This is the most general case and requires a more flexible model to capture the complex missingness patterns and infer the right target posterior distribution $p(\mathbf{X}_{\mis}\mid \mathbf{X}_{\obs}, \mathbf{R})$.  

In our setting, we allow for potentially non-ignorable (MNAR) patterns by explicitly modeling the missingness mechanism.  Hence we aim to learn a joint generative model of both the data-generating process and the missingness mechanism that yields a faithful approximation to the posterior $p(\mathbf{X}_{\mis}\mid \mathbf{X}_{\obs}, \mathbf{R})$.  Our goal is not only to accurately reconstruct the missing values, but to recover the full conditional distribution of the missing entries, providing both point imputations (e.g.\ posterior means) and uncertainty estimates (e.g.\ credible intervals).

\section{Method}

MissBGM is a Bayesian latent-variable model that jointly models the data-generating process and the missingness mechanism. The data-generating process is built upon the Bayesian generative modeling (BGM) framework~\citep{liu2026bayesian}, which has an established theoretical foundation and has been shown to be effective for modeling complex nonlinear dependencies in high-dimensional settings~\citep{Liu15042026,liu2021density}. MissBGM extends the BGM framework to missing data imputation by explicitly modeling the missingness mechanism together with the data-generating process. This allows the model to capture complex missingness patterns and infer the posterior distribution $p(\mathbf{X}_{\mis}\mid \mathbf{X}_{\obs}, \mathbf{R})$.

\subsection{Data and missingness generative process}

\begin{figure}[t]
    \centering
    \includegraphics[width=0.85\linewidth]{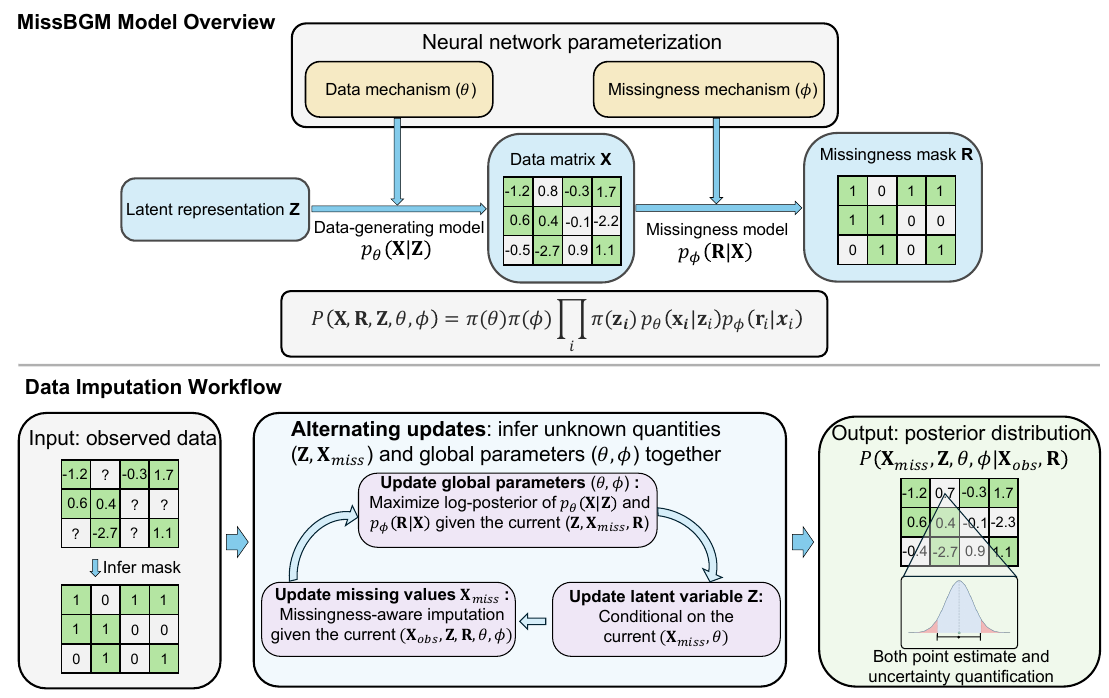}
    \caption{Overview of MissBGM. The model jointly represents a data-generating process $p_{\theta}(\mathbf{X}\mid \mathbf{Z})$ and a missingness mechanism $p_{\phi}(\mathbf{R}\mid \mathbf{X})$. During inference, observed entries remain fixed while latent variables and missing values are updated alternately. Per-sample posterior inference then provides both point estimates and uncertainty quantification for missing values.}
    \label{fig:overview}
\end{figure}

For each sample $i\in\{1,\ldots,n\}$, let $\mathbf{x}_i=({x}_{i1},\ldots,{x}_{ip})^\top \in \mathbb{R}^p$ denote the complete data vector and let $\mathbf{r}_i=({r}_{i1},\ldots,{r}_{ip})^\top \in \{0,1\}^p$ denote the corresponding observation mask. We introduce a latent variable $\mathbf{z}_i\in\mathbb{R}^d$ in the latent space and define the joint model
\begin{equation}
    p(\mathbf{X},\mathbf{R},\mathbf{Z},\theta,\phi)
    = \pi(\theta)\pi(\phi)\prod_{i=1}^n \pi(\mathbf{z}_i)\,p_{\theta}(\mathbf{x}_i\mid \mathbf{z}_i)\,p_{\phi}(\mathbf{r}_i\mid \mathbf{x}_i),
    \label{eq:joint}
\end{equation}
where $\pi(\mathbf{z}_i)=\mathcal{N}(0,I_d)$, $\pi(\theta)$ and $\pi(\phi)$ are prior distributions on the network parameters, $p_{\theta}(\mathbf{x}_i\mid \mathbf{z}_i)$ is the data model, and $p_{\phi}(\mathbf{r}_i\mid \mathbf{x}_i)$ is the missingness model. $d$ is the dimension of the latent variable. All the prior distributions are chosen to be multivariate normal distributions for simplicity.

By default, the data model $p_{\theta}(\mathbf{x}_i\mid \mathbf{z}_i)$ is set to be a multivariate normal distribution for continuous variables and a Bernoulli or Categorical distribution for discrete variables. In a typical continuous setting (e.g., tabular data), the data model is given by
\begin{equation}
    \mathbf{x}_i\mid \mathbf{z}_i,\theta \sim \mathcal{N}\!\left(\mu_{\theta}(\mathbf{z}_i),\Sigma_{\theta}(\mathbf{z}_i)\right),
    \label{eq:data_model}
\end{equation}
where both mean and covariance matrix are learnable functions of latent variable $\mathbf{z}_i$ parameterized by $\theta$. In practice, we can further simplify the covariance matrix to be a diagonal matrix $\Sigma_{\theta}(\mathbf{z}_i)=\operatorname{diag}(\sigma^2_{\theta,1}(\mathbf{z}_i),\ldots,\sigma^2_{\theta,p}(\mathbf{z}_i))$. We represent $\left(\mu_{\theta}(\mathbf{z}_i),\Sigma_{\theta}(\mathbf{z}_i)\right)$ as a neural network $g_{\theta}$ with two output heads, and the weights are shared across all samples.

We use another neural network $f_{\phi}(\mathbf{x}_i)\in\mathbb{R}^p$ to model the missingness mechanism and define
\begin{equation}
    r_{ij}\mid \mathbf{x}_i,\phi \sim \operatorname{Bernoulli}\!\left(\mathrm{sig}(f_{\phi,j}(\mathbf{x}_i))\right),
    \qquad j=1,\ldots,p,
    \label{eq:mask_model}
\end{equation}
where $\mathrm{sig}(x) = 1/(1 + \exp(-x))$. The probability of observing feature $j$ may depend on the full completed data vector $\mathbf{x}_i$, and hence the model can capture the comprehensive missingness process in a flexible and data-driven way.

To account for uncertainty in model parameter space, we can also adopt Bayesian neural networks~\citep{goan2020bayesian} for $g_{\theta}$ and $f_{\phi}$ to place priors on the network parameters. This gives a flexible nonparametric representation of both the data-generating mechanism and the missingness mechanism while preserving a coherent probabilistic model for uncertainty quantification.

\subsection{Alternating stochastic optimization}
We designed a stochastic optimization algorithm to update the model parameters, latent variables, and missing values iteratively until convergence. The joint posterior distribution is given by
\begin{equation}
    p(\mathbf{Z},\mathbf{X}_{\mis},\theta,\phi\mid \mathbf{X}_{\obs}, \mathbf{R})
    \propto \pi(\theta)\pi(\phi)\prod_{i=1}^n \pi(\mathbf{z}_i)\,p_{\theta}(\mathbf{x}_i\mid \mathbf{z}_i)\,p_{\phi}(\mathbf{r}_i\mid \mathbf{x}_i),
    \label{eq:full_posterior}
\end{equation}
where $\mathbf{X}=(\mathbf{X}_{\obs},\mathbf{X}_{\mis})$ is the complete data matrix. Since the true posterior joint distribution is not tractable, we approximate the target by the following iterative procedure. Specifically, we iteratively (i) update the latent variables $\mathbf{Z}$ from $p(\mathbf{Z}\mid \mathbf{X}, \mathbf{R}, \theta, \phi)$, (ii) update the missing values $\mathbf{X}_{\mis}$ from $p(\mathbf{X}_{\mis}\mid \mathbf{Z}, \mathbf{X}_{\obs}, \mathbf{R}, \theta, \phi)$, and (iii) update the model parameters $\theta$ and $\phi$ from $p(\theta, \phi\mid \mathbf{Z}, \mathbf{X}_{\mis}, \mathbf{X}_{\obs}, \mathbf{R})$.

\textbf{Step (i): update latent variable}. Up to an additive constant, the log-posterior is given by
\begin{equation}
    \log p(\mathbf{z}_i\mid \mathbf{x}_i, \mathbf{r}_i, \theta, \phi)
    = \log \pi(\mathbf{z}_i)+\log p_{\theta}(\mathbf{x}_i\mid \mathbf{z}_i)+C_1.
    \label{eq:z_posterior}
\end{equation}
Under diagonal covariance structure, the objective function is given by
\begin{equation}
    \mathcal{L}_{z_i}
    = \frac{1}{2}\lVert \mathbf{z}_i\rVert_2^2
    + \sum_{j=1}^p\left[
    \frac{(x_{ij}-\mu_{\theta,j}(\mathbf{z}_i))^2}{2\sigma^2_{\theta,j}(\mathbf{z}_i)}
    + \frac{1}{2}\log \sigma^2_{\theta,j}(\mathbf{z}_i)
    \right],
    \label{eq:z_objective}
\end{equation}
where $\mu_{\theta,j}(\mathbf{z}_i)$ and $\sigma^2_{\theta,j}(\mathbf{z}_i)$ are the $j$-th element of the mean and covariance vector, respectively. We update $\mathbf{z}_i$ by minimizing the objective function using gradient descent in \eqref{eq:z_objective} in parallel for all samples.

\textbf{Step (ii): update missing values}. Up to an additive constant, the log-posterior is given by
\begin{equation}
    \log p(\mathbf{x}_{i,\mis}\mid \mathbf{z}_i, \mathbf{x}_{i,\obs}, \mathbf{r}_i, \theta, \phi)
    = \log p_{\theta}(\mathbf{x}_i\mid \mathbf{z}_i)+\log p_{\phi}(\mathbf{r}_i\mid \mathbf{x}_i)+C_2.
    \label{eq:xmis_posterior}
\end{equation}
Under the diagonal Gaussian data model in \eqref{eq:data_model} and the Bernoulli missingness model in \eqref{eq:mask_model}, the objective function is given by
\begin{equation}
    \mathcal{L}_{x_{i,\mis}}
    = \sum_{j=1}^p \frac{(x_{ij}-\mu_{\theta,j}(\mathbf{z}_i))^2}{2\sigma^2_{\theta,j}(\mathbf{z}_i)}
    - \beta \sum_{j=1}^p \left[
    r_{ij}\log \mathrm{sig}(f_{\phi,j}(\mathbf{x}_i)) + (1-r_{ij})\log (1-\mathrm{sig}(f_{\phi,j}(\mathbf{x}_i)))
    \right],
    \label{eq:xmis_objective}
\end{equation}
where $\beta$ is a temperature coefficient to balance the data and missingness terms. We update $\mathbf{x}_{i,\mis}$ by minimizing the objective function using gradient descent in \eqref{eq:xmis_objective} in parallel for all samples.

\textbf{Step (iii): update model parameters}. Up to an additive constant, the log-posterior is given by
\begin{equation}
    \left\{
    \begin{aligned}
        \log p(\theta\mid \mathbf{Z}, \mathbf{X}) &= \log \pi(\theta)+\log p_{\theta}(\mathbf{X}\mid \mathbf{Z})+C_3, \\
        \log p(\phi\mid \mathbf{X}, \mathbf{R}) &= \log \pi(\phi)+\log p_{\phi}(\mathbf{R}\mid \mathbf{X})+C_4.
    \end{aligned}
    \right.
    \label{eq:param_logpost}
\end{equation}

To account for variation in model parameters with BNNs, which use variational inference (VI)~\citep{blei2017variational} to approximate \eqref{eq:param_logpost}, we can use the following evidence lower bounds (ELBOs) (see Appendix~\ref{app:elbo}):
\begin{equation}
    \left\{
    \begin{aligned}
        \mathcal{L}_{\theta}
        &= \mathbb{E}_{q_{\lambda}(\theta)}\!\left[\log p_{\theta}(\mathbf{X}\mid \mathbf{Z})\right]
        - \tau\,\mathrm{KL}\big(q_{\lambda}(\theta)\,\|\,\pi(\theta)\big), \\
        \mathcal{L}_{\phi}
        &= \mathbb{E}_{q_{\psi}(\phi)}\!\left[\log p_{\phi}(\mathbf{R}\mid \mathbf{X})\right]
        - \tau\,\mathrm{KL}\big(q_{\psi}(\phi)\,\|\,\pi(\phi)\big),
    \end{aligned}
    \right.
    \label{eq:elbo_bnn}
\end{equation}
where $\tau$ is the Kullback-Leibler divergence (KL) weight for both variational families. Two variational distributions $q_{\lambda}(\theta)=\mathcal{N}(\theta\mid \mu_{\lambda}, \sigma^2_{\lambda})$ and $q_{\psi}(\phi)=\mathcal{N}(\phi\mid \mu_{\psi}, \sigma^2_{\psi})$ are used to approximate the posterior distributions of $\theta$ and $\phi$, respectively. $\lambda=(\mu_{\lambda}, \sigma^2_{\lambda})$ and $\psi=(\mu_{\psi}, \sigma^2_{\psi})$ are the learnable variational parameters. Diagonal-covariance Gaussians are used for variantional distributions.

We update $\theta$ and $\phi$ by first maximizing the ELBOs with respect to $\lambda$ and $\psi$ in \eqref{eq:elbo_bnn} with stochastic gradients on mini-batches of samples, and then sampling $\theta$ and $\phi$ from the variational distributions $q_{\lambda}(\theta)$ and $q_{\psi}(\phi)$, respectively. 

After iterations of steps (i)-(iii) until convergence, we can obtain the point estimates of the missing values of the training dataset. The full alternating stochastic optimization algorithm is summarized in Algorithm~\ref{alg:missbgm-train}.

\subsection{Posterior inference and uncertainty quantification}\label{sec:posterior-inference}

Once the MissBGM model is trained, it performs posterior inference for incomplete samples without retraining the global network parameters. Given a (possibly new) test dataset with missingness pattern $\mathbf{R}$, our goal is to draw approximate samples from the full conditional posterior of the latent variables and missing values,
\begin{equation}
    p(\mathbf{Z},\mathbf{X}_{\mis}\mid \mathbf{X}_{\obs},\mathbf{R},\theta,\phi)
    \propto \prod_{i=1}^n \pi(\mathbf{z}_i)\,p_{\theta}(\mathbf{x}_i\mid \mathbf{z}_i)\,p_{\phi}(\mathbf{r}_i\mid \mathbf{x}_i),
    \label{eq:test_posterior}
\end{equation}
with the global parameters $(\theta,\phi)$ fixed (or drawn from fixed variational distributions $q_{\lambda}(\theta)$ and $q_{\psi}(\phi)$ in the BNN setting). All inference is per-sample, so the posterior sampling from \eqref{eq:test_posterior} can be done in parallel across samples to improve efficiency.

\textbf{Warm-start by MAP.}
We initialize $(\mathbf{Z},\mathbf{X}_{\mis})$ at a local maximum a posteriori (MAP) estimate by iterating only steps~(i) and (ii) of Algorithm~\ref{alg:missbgm-train} for a number of iterations $T_{\text{warm}}$. We denote the resulting warm-start state by $(\mathbf{Z}^{\star},\mathbf{X}_{\mis}^{\star})$, which already provides a good point estimate of the latent variables and missing values.

\textbf{HMC-within-Gibbs sampler.}
To sample from the full conditional posterior \eqref{eq:test_posterior}, we use the Hamiltonian Monte Carlo (HMC) within Gibbs sampler~\citep{tierney1994markov,neal2011mcmc}. Starting from $(\mathbf{Z}^{\star},\mathbf{X}_{\mis}^{\star})$, we draw posterior samples by alternating two HMC updates that share the same target \eqref{eq:test_posterior}:
\begin{equation}
    \left\{
    \begin{aligned}
        \log~&p(\mathbf{z}_i\mid \mathbf{x}_i,\theta) = \log~\pi(\mathbf{z}_i)+\log p_{\theta}(\mathbf{x}_i\mid \mathbf{z}_i)+C_5, \\
        \log~&p(\mathbf{x}_{i,\mis}\mid \mathbf{x}_{i,\obs},\mathbf{r}_i,\mathbf{z}_i,\theta,\phi) = \log~p_{\theta}(\mathbf{x}_i\mid \mathbf{z}_i)+\log~p_{\phi}(\mathbf{r}_i\mid \mathbf{x}_i)+C_6.
    \end{aligned}
    \right.
    \label{eq:hmc_targets}
\end{equation}
The two targets in \eqref{eq:hmc_targets} are exactly the negative per-sample objectives that drove the MAP warm-start in steps~(i) and (ii), so the sampler can be viewed as a stochastic refinement of MAP that further explores posterior uncertainty rather than a point estimate. At each Gibbs sweep, we (a) run one HMC step on $\mathbf{z}_i$ with $\mathbf{x}_{i}$ held fixed, then (b) run one HMC step on $\mathbf{x}_{i,\mis}$ with the just-updated $\mathbf{z}_i$ held fixed. Note that we use the same temperature coefficient $\beta$ as training stage.

Both HMC kernels use leapfrog integration with a small number of leapfrog steps and an adaptive step size controlled by a simple dual-averaging scheme targeting an acceptance rate of approximately $0.75$. Step-size adaptation is restricted to the first portion of the burn-in window and is then frozen, so that the post-burn-in draws form a valid Markov chain with respect to the fixed target. Since the targets in \eqref{eq:hmc_targets} are differentiable in both $\mathbf{z}_i$ and $\mathbf{x}_{i,\mis}$, gradient-based proposals mix substantially better than random-walk Metropolis, especially when $p$ is moderately large.

\textbf{Uncertainty quantification.} After discarding burn-in iterations, we collect $S$ joint draws $\{(\mathbf{Z}^{(s)},\mathbf{X}_{\mis}^{(s)})\}_{s=1}^{S}$ and reconstruct the corresponding completed matrices $\widetilde{\mathbf{X}}^{(s)}=\mathbf{R}\odot \mathbf{X}_{\obs}+(1-\mathbf{R})\odot \mathbf{X}_{\mis}^{(s)}$. From these posterior samples, we compute a posterior mean as a point estimate, and an empirical prediction interval (PI) with a user-specified significance level $\alpha$ from the lower and upper quantiles.
\begin{equation}
    \widehat{x}_{ij}=\frac{1}{S}\sum_{s=1}^{S} x_{ij}^{(s)},
    \qquad
    \widehat{\mathrm{PI}}_{ij}^{1-\alpha}=\Big[\,Q_{\alpha/2}\big(\{x_{ij}^{(s)}\}_{s=1}^{S}\big),\;Q_{1-\alpha/2}\big(\{x_{ij}^{(s)}\}_{s=1}^{S}\big)\,\Big],
    \label{eq:posterior_summary}
\end{equation}
where $Q_q(\cdot)$ denotes the $q$-quantile function. Unlike point-estimate imputers that only return $\widehat{\mathbf{X}}_{\mis}$, MissBGM thus delivers a full posterior of the missing values, providing much richer information for downstream tasks.

\section{Theoretical analysis}
The alternating stochastic updates in MissBGM optimize a tempered missing-data objective in \eqref{eq:xmis_objective} with the coefficient $\beta$. For one observed pair $(\mathbf{x}_{\obs},\mathbf{r})$, define
\begin{equation}
    m_{\beta}(\mathbf{x}_{\obs},\mathbf{r};\theta,\phi)
    :=
    \sup_{\mathbf{z},\mathbf{x}_{\mis}}
    \Big\{
    \log \pi(\mathbf{z}) + \log p_{\theta}(\mathbf{x}\mid \mathbf{z}) + \beta \log p_{\phi}(\mathbf{r}\mid \mathbf{x})
    \Big\},
    \qquad \mathbf{x}=(\mathbf{x}_{\obs},\mathbf{x}_{\mis}),
    \label{eq:theory_profiled_objective}
\end{equation}
and let $L_n^{\beta}(\theta,\phi)=n^{-1}\sum_{i=1}^n m_{\beta}(\mathbf{x}_{i,\obs},\mathbf{r}_i;\theta,\phi)$ with population counterpart $L^{\beta}(\theta,\phi)=\mathbb{E}_{P_0}[m_{\beta}(\mathbf{X}_{\obs},\mathbf{R};\theta,\phi)]$. For each $(\theta,\phi)$, let $q_{\theta,\phi}^{\beta}$ denote the normalized completed-data law proportional to $\int \pi(\mathbf{z})p_{\theta}(\mathbf{x}\mid \mathbf{z})p_{\phi}(\mathbf{r}\mid \mathbf{x})^{\beta}d\mathbf{z}$, and let $\hat p_{\theta,\phi}(\mathbf{x}_{\mis}\mid \mathbf{x}_{\obs},\mathbf{r})$ be its induced conditional law. This is the population analogue of the imputation distribution targeted by MissBGM.

\paragraph{Theorem 1 (Consistency).}
Assume the regularity conditions stated in Appendix~\ref{app:proof-theorem}: i.i.d. sampling, continuity and uniform convergence of the profiled criterion, vanishing sieve and optimization errors, and a separation condition ensuring that $L^{\beta}$ identifies a unique completed-data law $q_{\beta}^*$ on an evaluation set $\mathcal{S}$. If $(\hat\theta_n,\hat\phi_n)$ approximately maximizes $L_n^{\beta}$, then
\begin{equation}
    d_{\mathrm{jnt}}\bigl(q_{\hat\theta_n,\hat\phi_n}^{\beta},q_{\beta}^*\bigr)=o_p(1),
    \qquad
    \sup_{(\mathbf{x}_{\obs},\mathbf{r})\in\mathcal{S}}
    \bigl\|
    \hat p_{\hat\theta_n,\hat\phi_n}(\cdot\mid \mathbf{x}_{\obs},\mathbf{r})-p_{\beta}^*(\cdot\mid \mathbf{x}_{\obs},\mathbf{r})
    \bigr\|_{\mathrm{TV}}=o_p(1),
    \label{eq:main_theory_result}
\end{equation}
where $p_{\beta}^*$ is the conditional law induced by $q_{\beta}^*$ and $d_{\mathrm{jnt}}$ is the completed-data law distribution difference defined in Appendix~\ref{app:proof-theorem}. If, in addition, the support of $\mathbf{X}_{\mis}$ is uniformly bounded on $\mathcal{S}$, then the posterior-mean imputation rule $\hat\mu_n(\mathbf{x}_{\obs},\mathbf{r})=\int \mathbf{x}_{\mis}\,\hat p_{\hat\theta_n,\hat\phi_n}(d\mathbf{x}_{\mis}\mid \mathbf{x}_{\obs},\mathbf{r})$ satisfies
\begin{equation}
    \mathcal{R}_{\beta}^*(\hat\mu_n)-\mathcal{R}_{\beta}^*(\mu_{\beta}^*)=o_p(1),
    \label{eq:main_theory_risk}
\end{equation}
where $\mu_{\beta}^*$ is the pseudo-true conditional mean and $\mathcal{R}_{\beta}^*$ is the corresponding risk function.

The result shows that MissBGM is consistent for the \emph{pseudo-true} imputation law determined by the temperature parameter $\beta$. With $\beta=1$ and correct specification, $p_{\beta}^*$ reduces to the true conditional distribution $p(\mathbf{X}_{\mis}\mid \mathbf{X}_{\obs},\mathbf{R})$. Both the learned conditional law and posterior mean imputation converge to the corresponding Bayes-optimal limit. The full proof is given in Appendix~\ref{app:proof-theorem}.

\section{Experiments}

\subsection{Experimental setup}\label{sec:experimental-setup}

\textbf{Datasets.}
We evaluate MissBGM on a synthetic benchmark and four real-world tabular datasets that span a wide range of sample sizes and feature dimensions. The synthetic benchmark is designed such that the missing values follow a truncated Gaussian with a closed-form conditional distribution given the observed values, providing ground truth for both point imputation and uncertainty calibration. The real-data benchmark includes four commonly used datasets from the UCI Machine Learning Repository~\citep{asuncion2007uci} with different sample sizes ($n=178$ to $13,500$) and feature dimensions ($p=13$ to $5,000$). We add an MNAR missingness mechanism to each dataset with varying missing rates, where larger target coordinates are more likely to be missing. The full details of the datasets and missingness mechanism are given in Appendix~\ref{app:datasets}.

\textbf{Evaluation metrics.}
For point imputation, we report the root mean squared error (RMSE) on missing entries. For uncertainty quantification, we report the RMSE between per-entry estimated posterior standard deviations (S.D.) and the oracle S.D.s, the average width of the predicted $(1-\alpha)$ intervals, and the Pearson (PCC) and Spearman (SCC) correlations between predicted per-entry interval widths and oracle widths. Estimated posterior S.D.s are computed from $1{,}000$ multiple imputations (or posterior samples) per missing entry.

\textbf{Baseline methods.}

We compare MissBGM against seven data imputation baselines, including classical and nonparametric baselines (Mean), optimal-transport imputation (OT)~\citep{muzellec2020missing}, iterative chained equations (ICE)~\citep{van2011mice}, the ensemble multiple-imputation method (miceforest)~\citep{wilson_miceforest_2023}, the tree-based diffusion generative imputation method (ForestDiffusion)~\citep{jolicoeur2024generating}, and generative AI baselines (GAIN~\citep{yoon2018gain} and TabCSDI~\citep{zheng2022diffusion}). The full details of baselines implemented are given in Appendix~\ref{app:baselines}. The implementation details of MissBGM are given in Appendix~\ref{app:implementation}. The detailed instructions for using MissBGM software are given in Appendix~\ref{app:software-usage}.

\subsection{Synthetic experiments}\label{sec:synthetic}

\begin{table*}[t]
    \centering
    \caption{RMSE on synthetic datasets with varying sample sizes and missing rates. We report both mean and standard deviation of RMSE across 10 runs. Best result in each row is \textbf{bolded}.}
    \label{tab:synthetic-rmse}
    \setlength{\tabcolsep}{2.2pt}
    \renewcommand{\arraystretch}{0.9}
    \footnotesize
    \begin{tabular}{@{}cc cccccccc@{}}
        \toprule
        \textbf{Size} &
        \textbf{Rate} &
        \textbf{Mean}  & \textbf{OT} & \textbf{Miceforest} &
        \textbf{ICE} & \textbf{GAIN} &
        \textbf{ForestDiffusion} &
        \textbf{TabCSDI} & \textbf{MissBGM} \\
        \midrule
        \multirow{3}{*}{0.5k}
            & 0.3 & 1.849 & 1.664 (0.006) & 1.656 (0.011) & 1.652 (0.015) & 1.792 (0.018) & \textbf{1.617 (0.013)} & 1.665 (0.015) & 1.620 (0.011) \\
            & 0.5 & 1.817 & 1.640 (0.008) & 1.630 (0.014) & 1.611 (0.010) & 1.877 (0.014) & 1.641 (0.010) & 1.626 (0.010) & \textbf{1.585 (0.009)} \\
            & 0.7 & 1.887 & 1.728 (0.008) & 1.708 (0.012) & 1.696 (0.014) & 1.992 (0.015) & 1.795 (0.011) & 1.712 (0.013) & \textbf{1.673 (0.012)} \\
        \midrule
        \multirow{3}{*}{1k}
            & 0.3 & 1.798 & 1.602 (0.007) & 1.596 (0.011) & 1.591 (0.011) & 1.671 (0.017) & 1.575 (0.009) & 1.615 (0.010) & \textbf{1.568 (0.009)} \\
            & 0.5 & 1.765 & 1.574 (0.005) & 1.570 (0.013) & 1.556 (0.009) & 1.649 (0.012) & 1.567 (0.012) & 1.589 (0.009) & \textbf{1.538 (0.011)} \\
            & 0.7 & 1.838 & 1.674 (0.009) & 1.644 (0.014) & 1.636 (0.012) & 1.808 (0.015) & 1.727 (0.009) & 1.667 (0.012) & \textbf{1.619 (0.011)} \\
        \midrule
        \multirow{3}{*}{5k}
            & 0.3 & 1.816 & 1.632 (0.005) & 1.627 (0.010) & 1.618 (0.012) & 1.617 (0.013) & 1.629 (0.008) & 1.659 (0.010) & \textbf{1.599 (0.010)} \\
            & 0.5 & 1.793 & 1.603 (0.004) & 1.595 (0.009) & 1.586 (0.010) & 1.650 (0.010) & 1.599 (0.007) & 1.616 (0.009) & \textbf{1.574 (0.008)} \\
            & 0.7 & 1.870 & 1.697 (0.006) & 1.676 (0.011) & 1.666 (0.009) & 1.799 (0.012) & 1.736 (0.009) & 1.686 (0.010) & \textbf{1.662 (0.009)} \\
        \midrule
        \multirow{3}{*}{10k}
            & 0.3 & 1.772 & 1.592 (0.008) & 1.587 (0.008) & 1.577 (0.010) & 1.632 (0.011) & 1.608 (0.009) & 1.612 (0.007) & \textbf{1.562 (0.006)} \\
            & 0.5 & 1.750 & 1.566 (0.005) & 1.558 (0.007) & 1.548 (0.009) & \textbf{1.536 (0.009)} & 1.575 (0.008) & 1.575 (0.009) & 1.539 (0.005) \\
            & 0.7 & 1.824 & 1.661 (0.004) & 1.636 (0.009) & 1.625 (0.007) & 1.999 (0.012) & 1.728 (0.007) & 1.665 (0.010) & \textbf{1.624 (0.008)} \\
        \bottomrule
    \end{tabular}
    \vspace{-15pt}
\end{table*}

We first evaluate MissBGM on the synthetic benchmark by varying the sample size $n\in\{0.5\mathrm{k}, 1\mathrm{k}, 5\mathrm{k}, 10\mathrm{k}\}$ and the missing rate $r\in\{0.3, 0.5, 0.7\}$. $p=50$ for each configuration and the experiment is repeated for $10$ runs. As shown in Table~\ref{tab:synthetic-rmse}, MissBGM attains the lowest missing-entry RMSE in $10$ of the $12$ configurations. As expected, mean imputation is uniformly the weakest baseline across all settings. Classical conditional-modeling baselines (OT, miceforest, ICE) are competitive but consistently outperformed as they do not model the masking mechanism. The generative AI imputers, such as GAIN, ForestDiffusion, and TabCSDI, are competitive only in selected settings but still trailed by MissBGM in most settings. Overall, these results demonstrate that MissBGM provides robust and superior imputation accuracy across diverse settings.

\begin{wraptable}{r}{0.52\textwidth}
    %\vspace{-20pt}
    \vspace{-10pt}
    \centering
    \small
    \caption{Uncertainty quantification results on a synthetic dataset ($p=50$, $n=5000$, $\alpha=0.05$). The average oracle width is 1.539. Best in each column is \textbf{bolded}.}
    \label{tab:uq-sim}
    \setlength{\tabcolsep}{3.5pt}
    \renewcommand{\arraystretch}{1.05}
    \begin{tabular}{lcccc}
        \toprule
        \textbf{Method} & \textbf{RMSE (s.d.)} & \textbf{Avg.len} & \textbf{PCC} & \textbf{SCC} \\
        \midrule
        Miceforest        & 0.380 & \textbf{2.052} & 0.820 & 0.831 \\
        GAIN              & 0.933 & 0.008 & 0.275 & 0.297 \\
        ForestDiffusion   & 0.314   & 2.354   & \textbf{0.858}   & 0.851   \\
        TabCSDI           & 0.683   & 0.011   & 0.348   & 0.316   \\
        MissBGM           & \textbf{0.248} & 2.620 & 0.849 & \textbf{0.861} \\
        \bottomrule
    \end{tabular}
    \vspace{-10pt}
\end{wraptable}

Next, we evaluate the uncertainty quantification performance of MissBGM on a synthetic dataset against four baselines that provide multiple imputations. We use 1000 posterior samples for each missing value to compute the RMSE of standard deviation, interval width, and correlation between the predicted interval width and the oracle width for a user-specified confidence level $\alpha$. As shown in Table~\ref{tab:uq-sim}, although MissBGM obtains more conservative intervals than Miceforest and ForestDiffusion, it decreases RMSE in estimating the variation of the imputed values by $20.1\%$ compared to the strongest baseline ForestDiffusion. The interval-width correlations are also comparable to the best baseline. In contrast, the generative AI imputers (GAIN, TabCSDI) collapse to near-degenerate predictive distributions, yielding vanishing interval widths. This reflects a well-known failure mode of likelihood-free generative imputers, whose sample-to-sample variability severely underestimates posterior uncertainty.

\subsection{Real-world experiments}\label{sec:real-world}

\begin{wraptable}{r}{0.65\textwidth}
    %\vspace{-60pt}
    \vspace{-20pt}
    \centering
    \small
    \caption{RMSE on four real-world datasets with varying sample sizes and number of features. We report both mean and standard deviation of RMSE across 10 runs. Best result in each row is \textbf{bolded}.}
    \label{tab:real-rmse}
    \footnotesize
    \setlength{\tabcolsep}{3pt}
    \renewcommand{\arraystretch}{1.0}
    \begin{tabular}{@{}lcccc@{}}
        \toprule
        \textbf{Method} & \textbf{Wine} & \textbf{Breast} & \textbf{Concrete} & \textbf{Gisette} \\
        \midrule
        Mean              & 1.043  & 0.809 & 0.976  & 0.978  \\
        OT                & 0.697 (0.007) & 0.583 (0.005) & 0.538 (0.003) & 0.605 (0.004) \\
        Miceforest        & 0.698 (0.011) & 0.599 (0.008) & 0.556 (0.009) & - \\
        ICE               & 0.829 (0.014) & 0.618 (0.009) & 0.662 (0.012) & - \\
        GAIN              & 0.849 (0.017) & 0.651 (0.013) & 0.715 (0.015) & 0.696 (0.009) \\
        ForestDiffusion   & 0.779 (0.009) & 0.672 (0.006) & 0.655 (0.008) & - \\
        TabCSDI           & 0.691 (0.008) & 0.607 (0.006) & 0.569 (0.007) & 0.627 (0.004) \\
        MissBGM           & \textbf{0.682 (0.007)} & \textbf{0.511 (0.005)} & \textbf{0.529 (0.006)} & \textbf{0.543 (0.005)} \\
        \bottomrule
    \end{tabular}
    %\vspace{-30pt}
    \vspace{-10pt}
\end{wraptable}

We benchmark MissBGM against seven baselines on four standard tabular datasets in Table~\ref{tab:real-rmse}, spanning a wide range of sample sizes and feature dimensions, from Wine ($n=178$, $p=13$) to Gisette ($n=13{,}500$, $p=5{,}000$). MissBGM achieves the lowest mean RMSE across all four datasets. The advantage is substantial on the high-dimensional Gisette dataset, where MissBGM reduces RMSE by $10.2\%$ relative to the best competing method. Notably, three baselines, Miceforest, ICE, and ForestDiffusion fail to scale to Gisette (denoted by $-$). The results demonstrate that MissBGM scales well to high-dimensional data while maintaining superior imputation accuracy.

\subsection{Robustness and scalability}

\begin{figure}[t]
    \vspace{-15pt}
    \centering
    \includegraphics[width=0.85\linewidth]{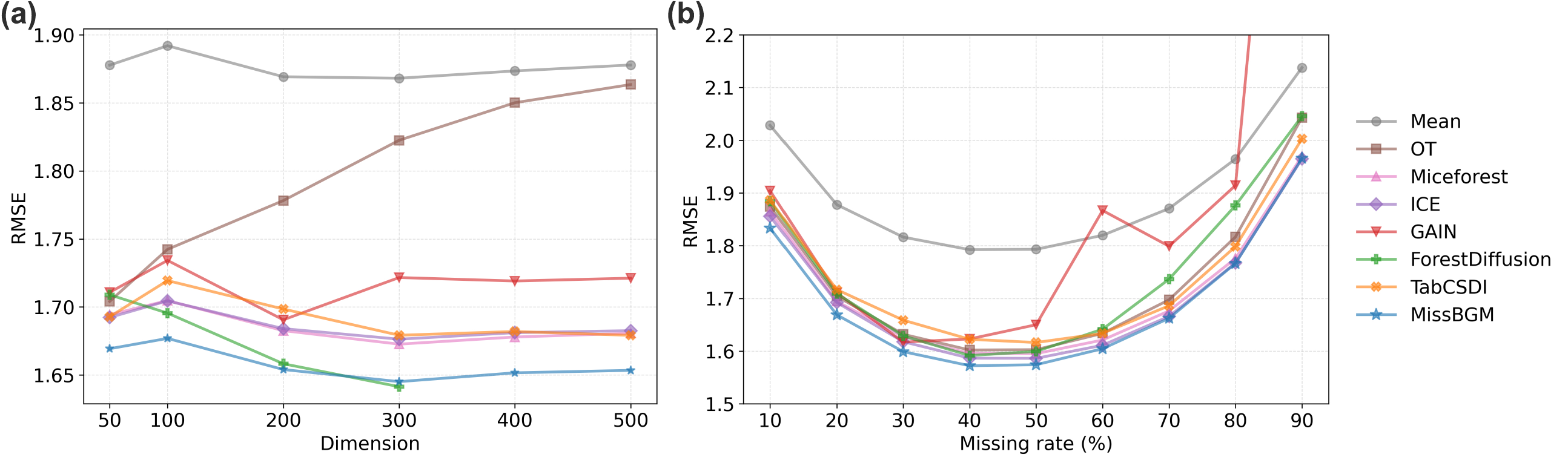}
    \caption{Robustness experiments on synthetic benchmark. (a) RMSE with varying dimensionality $p$ at $n=5000$ and missing rate $20\%$. (b) RMSE with varying missing rates at $n=5000$ and $p=50$.}
    \label{fig:robustness}
    \vspace{-15pt}
\end{figure}

We next assess the robustness of MissBGM regarding feature dimensionality and missingness severity (Figure~\ref{fig:robustness} and Appendix~\ref{app:missing-rate-size}). In Figure~\ref{fig:robustness}(a), OT degrades sharply with increasing $p$, reflecting the well-known difficulty of optimal transport in higher-dimensional settings. MissBGM achieves the lowest RMSE across all dimensions, except for $p=300$, where it is slightly outperformed by ForestDiffusion. In Figure~\ref{fig:robustness}(b), MissBGM achieves the lowest RMSE across the entire range, where competitors such as GAIN diverge and tree-based methods like ForestDiffusion deteriorate noticeably. We explain the non-monotonic results of missing rate in Section~\ref{app:datasets-synthetic}. These results suggest that MissBGM is robust to both dimensionality and missingness scale.

We further examine scalability from a computational perspective by comparing running time across methods on a moderate synthetic benchmark (Section~\ref{app:running-time}). The results show that MissBGM remains computationally practical and substantially faster than computationally intensive baselines such as miceforest and diffusion-based imputers, supporting its applicability to larger-scale problems.

\subsection{Ablation studies}

\begin{wraptable}{r}{0.45\textwidth}
    \vspace{-20pt}
    \centering
    \small
    \caption{Ablation study of MissBGM on the tempering coefficient $\beta$. RMSE on four real-world datasets. Best result in each row is \textbf{bolded}.}
    \label{tab:ablation-beta}
    \footnotesize
    \setlength{\tabcolsep}{4pt}
    \renewcommand{\arraystretch}{1.0}
    \begin{tabular}{@{}lcccc@{}}
        \toprule
        \textbf{Dataset} & $\beta=0$ & $\beta=0.01$ & $\beta=0.1$ & $\beta=1$ \\
        \midrule
        Wine     & 0.699          & \textbf{0.682} & 0.690          & 0.703 \\
        Breast   & 0.518          & 0.511          & \textbf{0.505} & 0.531 \\
        Concrete & \textbf{0.527} & 0.529          & 0.539          & 0.560 \\
        Gisette  & 0.557          & \textbf{0.543} & 0.553          & 0.564 \\
        \bottomrule
    \end{tabular}
    \vspace{-10pt}
\end{wraptable}

We study the role of the tempering coefficient $\beta$, which weights the missingness term in the tempered posterior of \eqref{eq:xmis_objective}. $\beta=0$ corresponds to ignoring the mask entirely. As shown in Table~\ref{tab:ablation-beta}, modeling the mask with a small but non-zero $\beta$ is beneficial in most scenarios. At the same time, performance degrades when $\beta$ becomes too large ($\beta=1$), suggesting that the mask should inform, but not dominate the imputation. Overall, we adopt $\beta=0.01$ across all experiments as the default. We also study the effect of using Bayesian neural networks (BNNs) on MissBGM (Appendix~\ref{app:bnn}).

\section{Conclusion}
We presented MissBGM, a missingness-aware Bayesian generative modeling approach for data imputation. MissBGM bridges the expressive flexibility of modern AI with the statistical rigor of Bayesian inference, enabling flexible representation learning with statistically rigorous imputation and uncertainty quantification. As with any joint model of data and missingness, reliability depends on appropriate specification of both the generative mechanism and the observation process. Neural networks training and Markov chain Monte Carlo (MCMC) can also be computationally intensive relative to simpler imputers. Future work includes (a) investigating identifiability of latent variables by leveraging recent advances in nonlinear independent component analysis (ICA) theory~\citep{khemakhem2020variational}, and (b) exploring more advanced architectures for $g_\theta$ and $f_\phi$ when the data dependence is more complex (e.g., non-stationary).

% \begin{ack}
% The authors would like to thank the reviewers for their helpful comments. The research presented in this paper was supported
% \end{ack}

\bibliographystyle{unsrtnat}
\bibliography{references}

\newpage
\appendix

\section{Derivation of ELBOs}\label{app:elbo}

This section derives the ELBO expressions in Eq.~\eqref{eq:elbo_bnn} from the conditional log-posteriors in Eq.~\eqref{eq:param_logpost}. Since the derivations for $\theta$ and $\phi$ are fully parallel, we present both in the same notation.

\paragraph{Step 1: conditional posteriors.}
From Eq.~\eqref{eq:param_logpost}, the conditional posteriors are, up to normalizing constants,
\begin{equation}
    p(\theta\mid \mathbf{Z},\mathbf{X})
    \propto p_{\theta}(\mathbf{X}\mid \mathbf{Z})\,\pi(\theta),
    \qquad
    p(\phi\mid \mathbf{X},\mathbf{R})
    \propto p_{\phi}(\mathbf{R}\mid \mathbf{X})\,\pi(\phi).
    \label{eq:appendix_conditional_posteriors}
\end{equation}
Equivalently, introducing the corresponding marginal likelihoods,
\begin{equation}
    p(\theta\mid \mathbf{Z},\mathbf{X})
    = \frac{p_{\theta}(\mathbf{X}\mid \mathbf{Z})\,\pi(\theta)}{p(\mathbf{X}\mid \mathbf{Z})},
    \qquad
    p(\phi\mid \mathbf{X},\mathbf{R})
    = \frac{p_{\phi}(\mathbf{R}\mid \mathbf{X})\,\pi(\phi)}{p(\mathbf{R}\mid \mathbf{X})},
    \label{eq:appendix_bayes_rule}
\end{equation}
where
\begin{equation}
    p(\mathbf{X}\mid \mathbf{Z}) = \int p_{\theta}(\mathbf{X}\mid \mathbf{Z})\,\pi(\theta)\,d\theta,
    \qquad
    p(\mathbf{R}\mid \mathbf{X}) = \int p_{\phi}(\mathbf{R}\mid \mathbf{X})\,\pi(\phi)\,d\phi.
    \label{eq:appendix_evidence_terms}
\end{equation}
These marginal likelihoods are generally intractable because they require integrating over all network parameters.

\paragraph{Step 2: variational approximation for $\theta$.}
Introduce a variational family $q_{\lambda}(\theta)$ to approximate $p(\theta\mid \mathbf{Z},\mathbf{X})$. Starting from the log evidence, we write
\begin{equation}
    \log p(\mathbf{X}\mid \mathbf{Z})
    = \log \int q_{\lambda}(\theta)
    \frac{p_{\theta}(\mathbf{X}\mid \mathbf{Z})\,\pi(\theta)}{q_{\lambda}(\theta)}\,d\theta.
    \label{eq:appendix_theta_insert_q}
\end{equation}
Applying Jensen's inequality gives
\begin{align}
    \log p(\mathbf{X}\mid \mathbf{Z})
    &\geq \int q_{\lambda}(\theta)
    \log \frac{p_{\theta}(\mathbf{X}\mid \mathbf{Z})\,\pi(\theta)}{q_{\lambda}(\theta)}\,d\theta \\
    &= \mathbb{E}_{q_{\lambda}(\theta)}\!\left[\log p_{\theta}(\mathbf{X}\mid \mathbf{Z})\right]
    + \mathbb{E}_{q_{\lambda}(\theta)}\!\left[\log \pi(\theta)\right]
    - \mathbb{E}_{q_{\lambda}(\theta)}\!\left[\log q_{\lambda}(\theta)\right].
    \label{eq:appendix_theta_jensen}
\end{align}
Using the definition of the KL divergence,
\begin{equation}
    \mathrm{KL}\big(q_{\lambda}(\theta)\,\|\,\pi(\theta)\big)
    = \mathbb{E}_{q_{\lambda}(\theta)}\!\left[\log q_{\lambda}(\theta) - \log \pi(\theta)\right],
    \label{eq:appendix_theta_kl_def}
\end{equation}
we obtain the standard ELBO
\begin{equation}
    \mathcal{L}^{\mathrm{std}}_{\theta}
    = \mathbb{E}_{q_{\lambda}(\theta)}\!\left[\log p_{\theta}(\mathbf{X}\mid \mathbf{Z})\right]
    - \mathrm{KL}\big(q_{\lambda}(\theta)\,\|\,\pi(\theta)\big),
    \label{eq:appendix_theta_elbo_standard}
\end{equation}
with
\begin{equation}
    \log p(\mathbf{X}\mid \mathbf{Z}) \geq \mathcal{L}^{\mathrm{std}}_{\theta}.
    \label{eq:appendix_theta_lower_bound}
\end{equation}
An equivalent decomposition, useful for interpretation, is
\begin{equation}
    \log p(\mathbf{X}\mid \mathbf{Z})
    = \mathcal{L}^{\mathrm{std}}_{\theta}
    + \mathrm{KL}\big(q_{\lambda}(\theta)\,\|\,p(\theta\mid \mathbf{Z},\mathbf{X})\big),
    \label{eq:appendix_theta_gap}
\end{equation}
which shows that maximizing the ELBO is equivalent to minimizing the KL divergence between the variational approximation and the exact conditional posterior.

\paragraph{Step 3: variational approximation for $\phi$.}
The same argument applies to the missingness-model parameters. Let $q_{\psi}(\phi)$ approximate $p(\phi\mid \mathbf{X},\mathbf{R})$. Then
\begin{equation}
    \log p(\mathbf{R}\mid \mathbf{X})
    = \log \int q_{\psi}(\phi)
    \frac{p_{\phi}(\mathbf{R}\mid \mathbf{X})\,\pi(\phi)}{q_{\psi}(\phi)}\,d\phi,
    \label{eq:appendix_phi_insert_q}
\end{equation}
so Jensen's inequality yields
\begin{align}
    \log p(\mathbf{R}\mid \mathbf{X})
    &\geq \mathbb{E}_{q_{\psi}(\phi)}\!\left[\log p_{\phi}(\mathbf{R}\mid \mathbf{X})\right]
    + \mathbb{E}_{q_{\psi}(\phi)}\!\left[\log \pi(\phi)\right]
    - \mathbb{E}_{q_{\psi}(\phi)}\!\left[\log q_{\psi}(\phi)\right] \\
    &= \mathbb{E}_{q_{\psi}(\phi)}\!\left[\log p_{\phi}(\mathbf{R}\mid \mathbf{X})\right]
    - \mathrm{KL}\big(q_{\psi}(\phi)\,\|\,\pi(\phi)\big).
    \label{eq:appendix_phi_elbo_standard}
\end{align}
Thus the standard ELBO for $\phi$ is
\begin{equation}
    \mathcal{L}^{\mathrm{std}}_{\phi}
    = \mathbb{E}_{q_{\psi}(\phi)}\!\left[\log p_{\phi}(\mathbf{R}\mid \mathbf{X})\right]
    - \mathrm{KL}\big(q_{\psi}(\phi)\,\|\,\pi(\phi)\big),
    \label{eq:appendix_phi_elbo_def}
\end{equation}
with the analogous decomposition
\begin{equation}
    \log p(\mathbf{R}\mid \mathbf{X})
    = \mathcal{L}^{\mathrm{std}}_{\phi}
    + \mathrm{KL}\big(q_{\psi}(\phi)\,\|\,p(\phi\mid \mathbf{X},\mathbf{R})\big).
    \label{eq:appendix_phi_gap}
\end{equation}

\paragraph{Step 4: weighted ELBOs used in MissBGM.}
The expressions in Eq.~\eqref{eq:elbo_bnn} are the KL-weighted versions of the standard ELBOs above:
\begin{equation}
    \left\{
    \begin{aligned}
        \mathcal{L}_{\theta}
        &= \mathbb{E}_{q_{\lambda}(\theta)}\!\left[\log p_{\theta}(\mathbf{X}\mid \mathbf{Z})\right]
        - \tau\,\mathrm{KL}\big(q_{\lambda}(\theta)\,\|\,\pi(\theta)\big), \\
        \mathcal{L}_{\phi}
        &= \mathbb{E}_{q_{\psi}(\phi)}\!\left[\log p_{\phi}(\mathbf{R}\mid \mathbf{X})\right]
        - \tau\,\mathrm{KL}\big(q_{\psi}(\phi)\,\|\,\pi(\phi)\big).
    \end{aligned}
    \right.
    \label{eq:appendix_weighted_elbo}
\end{equation}
When $\tau=0$, these reduce to the standard stochastic gradient descent updates. We set $\tau=0.00005$ as the default to control the strength of posterior regularization in Bayesian neural network training.

\paragraph{Step 5: Gaussian variational families.}
In MissBGM, we use Gaussian variational families
\begin{equation}
    q_{\lambda}(\theta)=\mathcal{N}(\theta\mid \mu_{\lambda},\sigma^2_{\lambda}I_{d_{\theta}}),
    \qquad
    q_{\psi}(\phi)=\mathcal{N}(\phi\mid \mu_{\psi},\sigma^2_{\psi}I_{d_{\phi}}),
    \label{eq:appendix_gaussian_variational_family}
\end{equation}
so the learnable quantities are the variational parameters $\lambda=(\mu_{\lambda},\sigma^2_{\lambda})$ and $\psi=(\mu_{\psi},\sigma^2_{\psi})$. Maximizing the objectives in Eq.~\eqref{eq:elbo_bnn} with respect to $\lambda$ and $\psi$ gives tractable stochastic updates for Bayesian neural-network training while preserving the decomposition into a data-fit term and a regularization term induced by the prior.

\section{Proof of theorem}\label{app:proof-theorem}
This section provides a detailed proof of Theorem~1. The key point is that MissBGM should be analyzed as a profiled $\beta$-tempered procedure, because the missing-value update in \eqref{eq:xmis_objective} optimizes the data term and the missingness term with the same temperature parameter $\beta$ that appears in the algorithm.

For notational convenience, write $O=(\mathbf{X}_{\obs},\mathbf{R})$ for a generic observed pair and let $P_0$ denote its data-generating distribution. For one realization $(\mathbf{x}_{\obs},\mathbf{r})$, define
\begin{equation}
    m_{\beta}(\mathbf{x}_{\obs},\mathbf{r};\theta,\phi)
    :=
    \sup_{\mathbf{z},\mathbf{x}_{\mis}}
    \Big\{
    \log \pi(\mathbf{z}) + \log p_{\theta}(\mathbf{x}\mid \mathbf{z}) + \beta \log p_{\phi}(\mathbf{r}\mid \mathbf{x})
    \Big\},
    \qquad \mathbf{x}=(\mathbf{x}_{\obs},\mathbf{x}_{\mis}),
    \label{eq:appendix_profiled_objective}
\end{equation}
with empirical and population criteria
\begin{equation}
    L_n^{\beta}(\theta,\phi)
    =
    \frac{1}{n}\sum_{i=1}^n m_{\beta}(\mathbf{x}_{i,\obs},\mathbf{r}_i;\theta,\phi),
    \qquad
    L^{\beta}(\theta,\phi)
    =
    \mathbb{E}_{P_0}\big[m_{\beta}(\mathbf{X}_{\obs},\mathbf{R};\theta,\phi)\big].
    \label{eq:appendix_empirical_population_objectives}
\end{equation}
To keep the asymptotic scaling standard, these criteria omit the $O(n^{-1})$ global prior terms $n^{-1}\log \pi(\theta)$ and $n^{-1}\log \pi(\phi)$. Adding them back does not affect any conclusion below.

For each $(\theta,\phi)$, let $q_{\theta,\phi}^{\beta}(\mathbf{x}_{\mis},\mathbf{x}_{\obs},\mathbf{r})$ denote the normalized completed-data law proportional to
\begin{equation}
    \int \pi(\mathbf{z})\,p_{\theta}(\mathbf{x}\mid \mathbf{z})\,p_{\phi}(\mathbf{r}\mid \mathbf{x})^{\beta}\,d\mathbf{z}.
    \label{eq:appendix_tempered_joint_law}
\end{equation}
Its induced conditional law is
\begin{equation}
    \hat p_{\theta,\phi}(\mathbf{x}_{\mis}\mid \mathbf{x}_{\obs},\mathbf{r})
    =
    \frac{q_{\theta,\phi}^{\beta}(\mathbf{x}_{\mis},\mathbf{x}_{\obs},\mathbf{r})}
    {\int q_{\theta,\phi}^{\beta}(\mathbf{u},\mathbf{x}_{\obs},\mathbf{r})\,d\mathbf{u}}.
    \label{eq:appendix_tempered_conditional_law}
\end{equation}
We now state the regularity conditions.

\paragraph{Assumption A1 (Sampling, measurability, and domination).}
The observed pairs $O_1,\ldots,O_n$ are i.i.d. from $P_0$. For every $(\theta,\phi)\in\Theta\times\Phi$, the map $O\mapsto m_{\beta}(O;\theta,\phi)$ is measurable and finite $P_0$-almost surely. There exists an integrable envelope $M_{\beta}$ such that
\begin{equation}
    \sup_{(\theta,\phi)\in\Theta\times\Phi}\bigl|m_{\beta}(O;\theta,\phi)\bigr|
    \le M_{\beta}(O)
    \quad P_0\text{-a.s.},
    \qquad
    \mathbb{E}_{P_0}\big[M_{\beta}(O)\big]<\infty.
    \label{eq:assumption_A1}
\end{equation}
The normalizing constant in \eqref{eq:appendix_tempered_joint_law} is finite for every $(\theta,\phi)$.

\paragraph{Assumption A2 (Compact sieves, continuity, and existence of a maximizer).}
There exist compact sieve sets $\Theta_n\subset\Theta$ and $\Phi_n\subset\Phi$ such that $\Theta_1\subset\Theta_2\subset\cdots$, $\Phi_1\subset\Phi_2\subset\cdots$, and $\overline{\cup_{n\ge 1}(\Theta_n\times\Phi_n)}=\Theta\times\Phi$. For $P_0$-almost every $O$, the map $(\theta,\phi)\mapsto m_{\beta}(O;\theta,\phi)$ is continuous on each $\Theta_n\times\Phi_n$, and $L^{\beta}$ extends continuously to a compactification on which its supremum is attained.

\paragraph{Assumption A3 (Uniform convergence of the profiled criterion).}
With
\begin{equation}
    \omega_n
    :=
    \sup_{(\theta,\phi)\in\Theta_n\times\Phi_n}
    \bigl|L_n^{\beta}(\theta,\phi)-L^{\beta}(\theta,\phi)\bigr|,
    \label{eq:omega_n_def}
\end{equation}
we have $\omega_n=O_p(a_n)$ for some deterministic sequence $a_n\downarrow0$.

\paragraph{Assumption A4 (Sieve approximation error).}
Define
\begin{equation}
    r_n
    :=
    \sup_{(\theta,\phi)\in\Theta\times\Phi}L^{\beta}(\theta,\phi)
    -
    \sup_{(\theta,\phi)\in\Theta_n\times\Phi_n}L^{\beta}(\theta,\phi).
    \label{eq:r_n_def}
\end{equation}
Then $r_n\to0$.

\paragraph{Assumption A5 (Vanishing optimization error).}
The fitted estimator $(\hat\theta_n,\hat\phi_n)\in\Theta_n\times\Phi_n$ satisfies
\begin{equation}
    \delta_n
    :=
    \sup_{(\theta,\phi)\in\Theta_n\times\Phi_n}L_n^{\beta}(\theta,\phi)
    -L_n^{\beta}(\hat\theta_n,\hat\phi_n)
    =O_p(b_n),
    \qquad b_n\downarrow0.
    \label{eq:delta_n_def}
\end{equation}

\paragraph{Assumption A6 (Pseudo-true target and law-level separation).}
The set
\begin{equation}
    \mathcal{M}_{\beta}^*
    :=
    \arg\max_{(\theta,\phi)\in\Theta\times\Phi}L^{\beta}(\theta,\phi)
    \label{eq:argmax_set_def}
\end{equation}
is nonempty. All elements of $\mathcal{M}_{\beta}^*$ induce the same completed-data law on an evaluation set $\mathcal{S}\subseteq\mathcal{X}_{\obs}\times\{0,1\}^p$; denote that law by $q_{\beta}^*$, and let $p_{\beta}^*(\mathbf{x}_{\mis}\mid \mathbf{x}_{\obs},\mathbf{r})$ be the corresponding conditional law. Define the metric
\begin{equation}
    d_{\mathrm{jnt}}(q_1,q_2)
    :=
    \sup_{(\mathbf{x}_{\obs},\mathbf{r})\in\mathcal{S}}
    \int
    \bigl|q_1(\mathbf{x}_{\mis},\mathbf{x}_{\obs},\mathbf{r})-q_2(\mathbf{x}_{\mis},\mathbf{x}_{\obs},\mathbf{r})\bigr|
    \,d\mathbf{x}_{\mis}.
    \label{eq:d_jnt_def}
\end{equation}
For every $\varepsilon>0$,
\begin{equation}
    \Delta_{\beta}(\varepsilon)
    :=
    \sup_{(\theta,\phi)\in\Theta\times\Phi}L^{\beta}(\theta,\phi)
    -
    \sup_{\{(\theta,\phi):\,d_{\mathrm{jnt}}(q_{\theta,\phi}^{\beta},q_{\beta}^*)\ge \varepsilon\}}
    L^{\beta}(\theta,\phi)
    >0.
    \label{eq:margin_function_def}
\end{equation}
Write
\begin{equation}
    \Delta_{\beta}^{\leftarrow}(t)
    :=
    \inf\bigl\{\varepsilon>0:\Delta_{\beta}(\varepsilon)\le t\bigr\}
    \label{eq:margin_inverse_def}
\end{equation}
for the generalized inverse of the separation margin.

\paragraph{Assumption A7 (Positivity of the observed-data margin).}
Let
\begin{equation}
    q_{\beta,\obs}^*(\mathbf{x}_{\obs},\mathbf{r})
    :=
    \int q_{\beta}^*(\mathbf{x}_{\mis},\mathbf{x}_{\obs},\mathbf{r})\,d\mathbf{x}_{\mis}.
    \label{eq:q_obs_star_def}
\end{equation}
There exists $c_0>0$ such that
\begin{equation}
    \inf_{(\mathbf{x}_{\obs},\mathbf{r})\in\mathcal{S}}q_{\beta,\obs}^*(\mathbf{x}_{\obs},\mathbf{r})\ge c_0.
    \label{eq:positive_margin_def}
\end{equation}
Hence $p_{\beta}^*(\cdot\mid \mathbf{x}_{\obs},\mathbf{r})$ is well defined on $\mathcal{S}$.

\paragraph{Assumption A8 (Bounded support for posterior means).}
There exists $B<\infty$ such that, for every $(\mathbf{x}_{\obs},\mathbf{r})\in\mathcal{S}$, the support of $\mathbf{X}_{\mis}$ under both $\hat p_{\theta,\phi}(\cdot\mid \mathbf{x}_{\obs},\mathbf{r})$ and $p_{\beta}^*(\cdot\mid \mathbf{x}_{\obs},\mathbf{r})$ is contained in a common Euclidean ball of radius $B$.

The assumption holds automatically if complete data live in a compact set (or after a fixed truncation of $\mathbf{X}_{\mis}$ to a large ball), and more generally encodes that, uniformly on $\mathcal{S}$, the relevant conditional mass for missing coordinates is not driven by unbounded tails in the objects we compare.

\paragraph{Theorem 1.}
Under Assumptions A1--A8,
\begin{equation}
    \sup_{(\theta,\phi)\in\Theta\times\Phi}L^{\beta}(\theta,\phi)-L^{\beta}(\hat\theta_n,\hat\phi_n)
    \le
    2\omega_n+r_n+\delta_n.
    \label{eq:restated_near_optimality}
\end{equation}
Consequently,
\begin{equation}
    d_{\mathrm{jnt}}\bigl(q_{\hat\theta_n,\hat\phi_n}^{\beta},q_{\beta}^*\bigr)
    =o_p(1),
    \qquad
    d_{\mathrm{jnt}}\bigl(q_{\hat\theta_n,\hat\phi_n}^{\beta},q_{\beta}^*\bigr)
    =O_p\Bigl(\Delta_{\beta}^{\leftarrow}(a_n+r_n+b_n)\Bigr),
    \label{eq:restated_joint_convergence}
\end{equation}
where the second statement uses the monotonicity of $\Delta_{\beta}^{\leftarrow}$ and the fact that $2\omega_n+r_n+\delta_n=O_p(a_n+r_n+b_n)$. Moreover,
\begin{equation}
    \sup_{(\mathbf{x}_{\obs},\mathbf{r})\in\mathcal{S}}
    \bigl\|
    \hat p_{\hat\theta_n,\hat\phi_n}(\cdot\mid \mathbf{x}_{\obs},\mathbf{r})-p_{\beta}^*(\cdot\mid \mathbf{x}_{\obs},\mathbf{r})
    \bigr\|_{\mathrm{TV}}
    \le
    \frac{2}{c_0}
    d_{\mathrm{jnt}}\bigl(q_{\hat\theta_n,\hat\phi_n}^{\beta},q_{\beta}^*\bigr)
    =o_p(1),
    \label{eq:restated_tv_bound}
\end{equation}
and if
\begin{equation}
    \hat\mu_n(\mathbf{x}_{\obs},\mathbf{r})
    :=
    \int \mathbf{x}_{\mis}\,\hat p_{\hat\theta_n,\hat\phi_n}(d\mathbf{x}_{\mis}\mid \mathbf{x}_{\obs},\mathbf{r}),
    \qquad
    \mu_{\beta}^*(\mathbf{x}_{\obs},\mathbf{r})
    :=
    \int \mathbf{x}_{\mis}\,p_{\beta}^*(d\mathbf{x}_{\mis}\mid \mathbf{x}_{\obs},\mathbf{r}),
    \label{eq:posterior_mean_rules}
\end{equation}
then the pseudo-true risk function
\begin{equation}
    \mathcal{R}_{\beta}^*(\mu)
    :=
    \mathbb{E}_{q_{\beta}^*}\big[\|\mathbf{X}_{\mis}-\mu(\mathbf{X}_{\obs},\mathbf{R})\|_2^2\big]
    \label{eq:pseudo_true_risk}
\end{equation}
satisfies
\begin{equation}
    0
    \le
    \mathcal{R}_{\beta}^*(\hat\mu_n)-\mathcal{R}_{\beta}^*(\mu_{\beta}^*)
    \le
    \frac{16B^2}{c_0^2}
    d_{\mathrm{jnt}}\bigl(q_{\hat\theta_n,\hat\phi_n}^{\beta},q_{\beta}^*\bigr)^2
    =o_p(1).
    \label{eq:restated_risk_bound}
\end{equation}
Under correct specification and $\beta=1$, the pseudo-true target equals the true conditional law $p(\mathbf{X}_{\mis}\mid \mathbf{X}_{\obs},\mathbf{R})$.

\paragraph{Lemma A1 (Population near-optimality of the fitted estimator).}
Under Assumptions A3--A5,
\begin{equation}
    \sup_{(\theta,\phi)\in\Theta\times\Phi}L^{\beta}(\theta,\phi)-L^{\beta}(\hat\theta_n,\hat\phi_n)
    \le
    2\omega_n+r_n+\delta_n.
    \label{eq:lemma_A1_claim}
\end{equation}

\begin{proof}
By the definition of $r_n$,
\[
    \sup_{(\theta,\phi)\in\Theta\times\Phi}L^{\beta}(\theta,\phi)
    \le
    \sup_{(\theta,\phi)\in\Theta_n\times\Phi_n}L^{\beta}(\theta,\phi)+r_n.
\]
By the definition of $\omega_n$,
\[
    \sup_{(\theta,\phi)\in\Theta_n\times\Phi_n}L^{\beta}(\theta,\phi)
    \le
    \sup_{(\theta,\phi)\in\Theta_n\times\Phi_n}L_n^{\beta}(\theta,\phi)+\omega_n.
\]
By the definition of $\delta_n$,
\[
    \sup_{(\theta,\phi)\in\Theta_n\times\Phi_n}L_n^{\beta}(\theta,\phi)
    \le
    L_n^{\beta}(\hat\theta_n,\hat\phi_n)+\delta_n.
\]
Applying the definition of $\omega_n$ once more,
\[
    L_n^{\beta}(\hat\theta_n,\hat\phi_n)
    \le
    L^{\beta}(\hat\theta_n,\hat\phi_n)+\omega_n.
\]
Combining the four displays gives \eqref{eq:lemma_A1_claim}.
\end{proof}

\paragraph{Lemma A2 (Joint-law consistency).}
Under Assumptions A3--A6,
\begin{equation}
    d_{\mathrm{jnt}}\bigl(q_{\hat\theta_n,\hat\phi_n}^{\beta},q_{\beta}^*\bigr)=o_p(1),
    \qquad
    d_{\mathrm{jnt}}\bigl(q_{\hat\theta_n,\hat\phi_n}^{\beta},q_{\beta}^*\bigr)
    =O_p\Bigl(\Delta_{\beta}^{\leftarrow}(2\omega_n+r_n+\delta_n)\Bigr).
    \label{eq:lemma_A2_claim}
\end{equation}

\begin{proof}
Fix $\varepsilon>0$. On the event
\[
    \bigl\{d_{\mathrm{jnt}}(q_{\hat\theta_n,\hat\phi_n}^{\beta},q_{\beta}^*)\ge \varepsilon\bigr\},
\]
Assumption A6 implies
\[
    L^{\beta}(\hat\theta_n,\hat\phi_n)
    \le
    \sup_{(\theta,\phi)\in\Theta\times\Phi}L^{\beta}(\theta,\phi)-\Delta_{\beta}(\varepsilon).
\]
Equivalently,
\[
    \sup_{(\theta,\phi)\in\Theta\times\Phi}L^{\beta}(\theta,\phi)-L^{\beta}(\hat\theta_n,\hat\phi_n)
    \ge
    \Delta_{\beta}(\varepsilon).
\]
Therefore, by Lemma A1,
\[
    \mathbb{P}\Bigl(
    d_{\mathrm{jnt}}(q_{\hat\theta_n,\hat\phi_n}^{\beta},q_{\beta}^*)\ge \varepsilon
    \Bigr)
    \le
    \mathbb{P}\bigl(2\omega_n+r_n+\delta_n\ge \Delta_{\beta}(\varepsilon)\bigr).
\]
Since $\omega_n=O_p(a_n)$, $r_n\to0$, and $\delta_n=O_p(b_n)$, the right-hand side converges to zero for every fixed $\varepsilon>0$, proving the first claim.

For the rate statement, let $t_n:=2\omega_n+r_n+\delta_n$. The previous argument yields
\[
    \{d_{\mathrm{jnt}}(q_{\hat\theta_n,\hat\phi_n}^{\beta},q_{\beta}^*)>\varepsilon\}
    \subseteq
    \{t_n\ge \Delta_{\beta}(\varepsilon)\}.
\]
By the definition of the generalized inverse in \eqref{eq:margin_inverse_def}, this implies
\[
    d_{\mathrm{jnt}}(q_{\hat\theta_n,\hat\phi_n}^{\beta},q_{\beta}^*)
    \le
    \Delta_{\beta}^{\leftarrow}(t_n)
\]
with probability tending to one, which is equivalent to the second claim in \eqref{eq:lemma_A2_claim}.
\end{proof}

\paragraph{Lemma A3 (Conditional-law consistency from completed-data-law consistency).}
Under Assumptions A6 and A7,
\begin{equation}
    \sup_{(\mathbf{x}_{\obs},\mathbf{r})\in\mathcal{S}}
    \bigl\|
    \hat p_{\hat\theta_n,\hat\phi_n}(\cdot\mid \mathbf{x}_{\obs},\mathbf{r})-p_{\beta}^*(\cdot\mid \mathbf{x}_{\obs},\mathbf{r})
    \bigr\|_{\mathrm{TV}}
    \le
    \frac{2}{c_0}
    d_{\mathrm{jnt}}\bigl(q_{\hat\theta_n,\hat\phi_n}^{\beta},q_{\beta}^*\bigr)
    \label{eq:lemma_A3_claim}
\end{equation}
whenever $d_{\mathrm{jnt}}(q_{\hat\theta_n,\hat\phi_n}^{\beta},q_{\beta}^*)\le c_0/2$.

\begin{proof}
Fix $(\mathbf{x}_{\obs},\mathbf{r})\in\mathcal{S}$ and write
\[
    q_n(\mathbf{u})
    :=
    q_{\hat\theta_n,\hat\phi_n}^{\beta}(\mathbf{u},\mathbf{x}_{\obs},\mathbf{r}),
    \qquad
    q_*(\mathbf{u})
    :=
    q_{\beta}^*(\mathbf{u},\mathbf{x}_{\obs},\mathbf{r}).
\]
Let
\[
    A_n:=\int q_n(\mathbf{u})\,d\mathbf{u},
    \qquad
    A_*:=\int q_*(\mathbf{u})\,d\mathbf{u}=q_{\beta,\obs}^*(\mathbf{x}_{\obs},\mathbf{r}).
\]
By Assumption A7, $A_*\ge c_0$. Moreover,
\[
    |A_n-A_*|
    \le
    \int |q_n(\mathbf{u})-q_*(\mathbf{u})|\,d\mathbf{u}
    \le
    d_{\mathrm{jnt}}\bigl(q_{\hat\theta_n,\hat\phi_n}^{\beta},q_{\beta}^*\bigr).
\]
Hence if $d_{\mathrm{jnt}}(q_{\hat\theta_n,\hat\phi_n}^{\beta},q_{\beta}^*)\le c_0/2$, then $A_n\ge c_0/2$.

Now use
\[
    \hat p_{\hat\theta_n,\hat\phi_n}(\mathbf{u}\mid \mathbf{x}_{\obs},\mathbf{r})=\frac{q_n(\mathbf{u})}{A_n},
    \qquad
    p_{\beta}^*(\mathbf{u}\mid \mathbf{x}_{\obs},\mathbf{r})=\frac{q_*(\mathbf{u})}{A_*}.
\]
Then
\[
    \left\|\frac{q_n}{A_n}-\frac{q_*}{A_*}\right\|_1
    \le
    \frac{\|q_n-q_*\|_1}{A_n}
    +
    \left|\frac{1}{A_n}-\frac{1}{A_*}\right|\|q_*\|_1.
\]
Since $\|q_*\|_1=A_*$,
\[
    \left\|\frac{q_n}{A_n}-\frac{q_*}{A_*}\right\|_1
    \le
    \frac{\|q_n-q_*\|_1}{A_n}
    +
    \frac{|A_n-A_*|}{A_n}
    \le
    \frac{2\|q_n-q_*\|_1}{A_n}
    \le
    \frac{4}{c_0}\|q_n-q_*\|_1.
\]
Dividing by $2$ converts $L^1$ distance to total variation distance, so
\[
    \bigl\|
    \hat p_{\hat\theta_n,\hat\phi_n}(\cdot\mid \mathbf{x}_{\obs},\mathbf{r})-p_{\beta}^*(\cdot\mid \mathbf{x}_{\obs},\mathbf{r})
    \bigr\|_{\mathrm{TV}}
    \le
    \frac{2}{c_0}\|q_n-q_*\|_1.
\]
Taking the supremum over $(\mathbf{x}_{\obs},\mathbf{r})\in\mathcal{S}$ proves \eqref{eq:lemma_A3_claim}.
\end{proof}

\paragraph{Lemma A4 (Excess-risk bound for posterior-mean imputation).}
Under Assumption A8,
\begin{equation}
    0
    \le
    \mathcal{R}_{\beta}^*(\hat\mu_n)-\mathcal{R}_{\beta}^*(\mu_{\beta}^*)
    \le
    \frac{16B^2}{c_0^2}
    d_{\mathrm{jnt}}\bigl(q_{\hat\theta_n,\hat\phi_n}^{\beta},q_{\beta}^*\bigr)^2.
    \label{eq:lemma_A4_claim}
\end{equation}

\begin{proof}
For every $(\mathbf{x}_{\obs},\mathbf{r})\in\mathcal{S}$,
\[
    \|\hat\mu_n(\mathbf{x}_{\obs},\mathbf{r})-\mu_{\beta}^*(\mathbf{x}_{\obs},\mathbf{r})\|_2
    =
    \left\|
    \int \mathbf{u}\,
    \bigl[\hat p_{\hat\theta_n,\hat\phi_n}(d\mathbf{u}\mid \mathbf{x}_{\obs},\mathbf{r})-p_{\beta}^*(d\mathbf{u}\mid \mathbf{x}_{\obs},\mathbf{r})\bigr]
    \right\|_2.
\]
Because the support of $\mathbf{u}$ is contained in a ball of radius $B$,
\[
    \|\hat\mu_n(\mathbf{x}_{\obs},\mathbf{r})-\mu_{\beta}^*(\mathbf{x}_{\obs},\mathbf{r})\|_2
    \le
    B\,\bigl\|\hat p_{\hat\theta_n,\hat\phi_n}(\cdot\mid \mathbf{x}_{\obs},\mathbf{r})-p_{\beta}^*(\cdot\mid \mathbf{x}_{\obs},\mathbf{r})\bigr\|_1.
\]
Applying Lemma A3 gives
\[
    \|\hat\mu_n(\mathbf{x}_{\obs},\mathbf{r})-\mu_{\beta}^*(\mathbf{x}_{\obs},\mathbf{r})\|_2
    \le
    \frac{4B}{c_0}
    d_{\mathrm{jnt}}\bigl(q_{\hat\theta_n,\hat\phi_n}^{\beta},q_{\beta}^*\bigr).
\]
Squaring both sides yields the pointwise bound
\[
    \|\hat\mu_n(\mathbf{x}_{\obs},\mathbf{r})-\mu_{\beta}^*(\mathbf{x}_{\obs},\mathbf{r})\|_2^2
    \le
    \frac{16B^2}{c_0^2}
    d_{\mathrm{jnt}}\bigl(q_{\hat\theta_n,\hat\phi_n}^{\beta},q_{\beta}^*\bigr)^2.
\]
Under $q_{\beta}^*$, the conditional expectation of $\mathbf{X}_{\mis}$ given $(\mathbf{X}_{\obs},\mathbf{R})$ is $\mu_{\beta}^*(\mathbf{X}_{\obs},\mathbf{R})$. Therefore the standard conditional-mean orthogonality identity implies
\[
    \mathcal{R}_{\beta}^*(\hat\mu_n)-\mathcal{R}_{\beta}^*(\mu_{\beta}^*)
    =
    \mathbb{E}_{q_{\beta}^*}\big[\|\hat\mu_n(\mathbf{X}_{\obs},\mathbf{R})-\mu_{\beta}^*(\mathbf{X}_{\obs},\mathbf{R})\|_2^2\big].
\]
Taking expectations in the previous pointwise inequality proves \eqref{eq:lemma_A4_claim}. Nonnegativity is immediate from the Bayes optimality of the conditional mean under squared loss.
\end{proof}

\begin{proof}[Proof of Theorem~1]
The near-optimality bound \eqref{eq:restated_near_optimality} is exactly Lemma A1. Lemma A2 then yields convergence of the completed-data law to the pseudo-true tempered target, establishing \eqref{eq:restated_joint_convergence}. Lemma A3 converts that completed-data-law convergence into the total-variation bound \eqref{eq:restated_tv_bound} for the induced conditional imputation law. Finally, Lemma A4 gives the excess-risk bound \eqref{eq:restated_risk_bound} for the posterior-mean imputation rule.

When the model is correctly specified and $\beta=1$, the profiled population criterion is maximized at the true completed-data law, so the pseudo-true target $p_{\beta}^*$ coincides with the genuine conditional law $p(\mathbf{X}_{\mis}\mid \mathbf{X}_{\obs},\mathbf{R})$. In that case Theorem~1 reduces to consistency for both the full imputation distribution and the induced posterior-mean estimator.
\end{proof}

The current theorem is stated with deterministic fitted global parameters $(\hat\theta_n,\hat\phi_n)$. For Bayesian neural network variant, we can replace $(\theta,\phi)$ by the variational parameters and absorb the ELBO approximation error into the analogue of $\delta_n$. The structural conclusion will remain the same: consistency is with respect to the pseudo-true target defined by the same tempered objective that MissBGM actually optimizes.

\section{Datasets}\label{app:datasets}

We evaluate MissBGM on a synthetic benchmark and four real-world tabular datasets that span a wide range of sample sizes and feature dimensions. This appendix details the data-generating processes, the missingness mechanisms, and the preprocessing protocol shared across all methods.

\subsection{Synthetic dataset}\label{app:datasets-synthetic}

The synthetic benchmark is generated through a self-masked Gaussian model that yields closed-form ground-truth conditional means, variances, and prediction intervals for every missing entry, which makes it an ideal benchmark for both point imputation and uncertainty calibration. The synthetic benchmark is generated with the following process.

\paragraph{Anchor and target partition.}
Each sample $\mathbf{x}_i\in\mathbb{R}^p$ is partitioned into a fully observed \emph{anchor} block of size $p_a$ and a potentially missing \emph{target} block of size $p_t=p-p_a$. We index the two blocks by the disjoint coordinate sets
\begin{equation}
    \mathcal{J}_a=\{1,\ldots,p_a\},\qquad \mathcal{J}_t=\{p_a+1,\ldots,p\},
\end{equation}
and write the corresponding subvectors as $\mathbf{x}_{i,\mathcal{J}_a}=(x_{ij})_{j\in\mathcal{J}_a}\in\mathbb{R}^{p_a}$ and $\mathbf{x}_{i,\mathcal{J}_t}=(x_{ij})_{j\in\mathcal{J}_t}\in\mathbb{R}^{p_t}$. The full sample is then assembled by concatenation,
\begin{equation}
    \mathbf{x}_i=\big(\mathbf{x}_{i,\mathcal{J}_a},\,\mathbf{x}_{i,\mathcal{J}_t}\big)\in\mathbb{R}^p.
\end{equation}
Throughout the synthetic experiments we set $p_a=5$, so that $p_t=p-5$.

\paragraph{Global parameters.}
The model has three population-level parameters $(B,\mathbf{b},\boldsymbol{\sigma})$ that are sampled \emph{once} per simulation replicate from a fixed seed and then shared across all $n$ data points,
\begin{equation}
    B\in\mathbb{R}^{p_a\times p_t},\;\; B_{kl}\stackrel{\mathrm{iid}}{\sim}\mathcal{N}(0,\,0.4^2);\quad
    \mathbf{b}\in\mathbb{R}^{p_t},\;\; b_l\stackrel{\mathrm{iid}}{\sim}\mathcal{N}(0,\,0.3^2);\quad
    \sigma_l\stackrel{\mathrm{iid}}{\sim}\mathrm{Uniform}(0.6,\,1.2),
    \label{eq:oracle_global_params}
\end{equation}
for $k=1,\ldots,p_a$ and $l=1,\ldots,p_t$, with $\boldsymbol{\sigma}=(\sigma_1,\ldots,\sigma_{p_t})^{\top}$. To match the per-target-coordinate notation used below, we define $\sigma_{(j)}\equiv \sigma_{j-p_a}$ and $b_{(j)}\equiv b_{j-p_a}$ for any target index $j\in\mathcal{J}_t$, and let $B_{:,(j)}\in\mathbb{R}^{p_a}$ denote the column of $B$ corresponding to target coordinate $j$.

\paragraph{Per-sample data generation.}
Conditional on $(B,\mathbf{b},\boldsymbol{\sigma})$, each of the $n$ samples is generated independently as
\begin{equation}
    \mathbf{x}_{i,\mathcal{J}_a}\sim \mathcal{N}(\mathbf{0},\,I_{p_a}),\qquad
    \mathbf{x}_{i,\mathcal{J}_t}\mid \mathbf{x}_{i,\mathcal{J}_a}\sim \mathcal{N}\!\left(B^{\top}\mathbf{x}_{i,\mathcal{J}_a}+\mathbf{b},\,\mathrm{diag}(\boldsymbol{\sigma}^2)\right),
    \label{eq:oracle_data_model}
\end{equation}
where $\mathrm{diag}(\boldsymbol{\sigma}^2)\in\mathbb{R}^{p_t\times p_t}$ is the diagonal matrix with entries $\sigma_1^2,\ldots,\sigma_{p_t}^2$. Equivalently, defining
\begin{equation}
    \mu_j(\mathbf{x}_{i,\mathcal{J}_a})\;=\;B_{:,(j)}^{\top}\mathbf{x}_{i,\mathcal{J}_a}+b_{(j)},\qquad j\in\mathcal{J}_t,
\end{equation}
the marginal conditional law of each target coordinate is the heteroscedastic Gaussian
\begin{equation}
    x_{ij}\mid \mathbf{x}_{i,\mathcal{J}_a}\;\sim\;\mathcal{N}\!\left(\mu_j(\mathbf{x}_{i,\mathcal{J}_a}),\,\sigma_{(j)}^2\right),\qquad j\in\mathcal{J}_t.
    \label{eq:oracle_marginal}
\end{equation}

\paragraph{Self-masked MNAR rule.}
Anchor coordinates are always observed: $r_{ij}=1$ for all $j\in\mathcal{J}_a$. For each target coordinate $j\in\mathcal{J}_t$, the entry is observed if and only if its realization falls below a quantile threshold of its own conditional distribution,
\begin{equation}
    r_{ij}=\mathbb{I}\!\left\{x_{ij}\le \mu_{j}(\mathbf{x}_{i,\mathcal{J}_a})+\sigma_{(j)}\kappa\right\},\qquad \kappa=\Phi^{-1}(1-r),
    \label{eq:oracle_mask}
\end{equation}
where $\Phi$ is the standard normal cumulative distribution function (CDF) and $r\in(0,1)$ is the targeted per-column missing rate on the target block. Because the threshold depends on the value of $x_{ij}$ itself, larger realizations of $x_{ij}$ are systematically more likely to be missing, so the masking mechanism is non-ignorable (MNAR). 

\paragraph{Effect of missing rate $r$.}
This self-masked construction also explains why the RMSE in Figure~\ref{fig:robustness}(b) is non-monotonic in $r$. When $r$ is small, $\kappa=\Phi^{-1}(1-r)$ is large, so only the most extreme right-tail realizations of $x_{ij}$ are masked; these atypical values lie far from the conditional mean $\mu_j(\mathbf{x}_{i,\mathcal{J}_a})$ and are intrinsically hard to recover even when abundant observed entries are available for training. When $r$ is large, the threshold falls into the bulk of the conditional distribution and a large fraction of target entries are missing, leaving too few observations to fit the conditional model reliably. The hardest regimes therefore arise at both ends of the missing-rate spectrum, while moderate missing rates strike a balance between informativeness of the observed entries and severity of the tail truncation, which yields the lowest imputation RMSE in Figure~\ref{fig:robustness}(b).

\paragraph{Oracle posterior and prediction interval.}
Conditional on being missing, $x_{ij}$ follows a Gaussian truncated from above at $\mu_j(\mathbf{x}_{i,\mathcal{J}_a})+\sigma_{(j)}\kappa$, so its posterior quantiles are available in closed form. Letting $\mu_j=\mu_j(\mathbf{x}_{i,\mathcal{J}_a})$ and $\Phi_\kappa=\Phi(\kappa)$, the oracle $(1-\alpha)$ prediction interval (PI) for a missing entry $(i,j)$ with $j\in\mathcal{J}_t$ is
\begin{equation}
    \mathrm{PI}_{ij}^{1-\alpha,\,\star}=\Big[\,\mu_j+\sigma_{(j)}\Phi^{-1}(p_{\mathrm{lo}}),\;\mu_j+\sigma_{(j)}\Phi^{-1}(p_{\mathrm{hi}})\,\Big],
    \label{eq:oracle_interval}
\end{equation}
with lower and upper quantile levels
\begin{equation}
    p_{\mathrm{lo}}=\Phi_\kappa+\tfrac{\alpha}{2}\big(1-\Phi_\kappa\big),\qquad
    p_{\mathrm{hi}}=\Phi_\kappa+\big(1-\tfrac{\alpha}{2}\big)\big(1-\Phi_\kappa\big).
\end{equation}
The oracle point imputation is the truncated-Gaussian mean, and the oracle per-entry interval width used in the PCC and SCC metrics is simply the length of $\mathrm{PI}_{ij}^{1-\alpha,\,\star}$ given a user-specified significance level $\alpha$. We use $\alpha=0.05$ as default in all synthetic experiments.

\subsection{Real-world datasets}\label{app:datasets-real}

We use four publicly available tabular datasets from the UCI Machine Learning Repository (\url{https://archive.ics.uci.edu/}) as our real-world benchmark. They vary in sample size and feature dimension from small-$n$ low-dimensional regression problems to a very high-dimensional dataset. A summary of the datasets is given in Table~\ref{tab:real-datasets}.

\begin{table}[h]
    \centering
    \small
    \caption{Summary of the four real-world tabular datasets used in our benchmark. All datasets are obtained from the UCI Machine Learning Repository. The reported \#Samples and \#Features are after dropping ID columns, target columns, and rows with native missing values where applicable.}
    \label{tab:real-datasets}
    \setlength{\tabcolsep}{6pt}
    \renewcommand{\arraystretch}{1.1}
    \begin{tabular}{@{}lcccc@{}}
        \toprule
        \textbf{Dataset} & \textbf{UCI ID} & \textbf{\#Samples} & \textbf{\#Features} & \textbf{Domain} \\
        \midrule
        Wine     & 109 & 178   & 13   & Chemistry / agriculture \\
        Concrete & 165 & 1{,}030 & 8    & Civil engineering        \\
        Breast   & 15  & 683   & 9    & Biomedical               \\
        Gisette  & 170 & 13{,}500 & 5{,}000 & Computer vision (digits) \\
        \bottomrule
    \end{tabular}
\end{table}

\textbf{Wine} contains $n=178$ chemical analyses of wines grown in the same region in Italy but derived from three different cultivars. We drop the class label column and use the remaining $p=13$ continuous chemical attributes (alcohol, malic acid, flavanoids, color intensity, etc.) as the imputation target. 

\textbf{Concrete Compressive Strength} contains $n=1,030$ concrete mixture records with $p=8$ continuous predictors describing the mixture composition (cement, slag, fly ash, water, superplasticizer, coarse and fine aggregates, age). We drop the target compressive strength column and treat the remaining columns as features. 

\textbf{Breast Cancer Wisconsin (Original)} contains digitized cytology measurements with $p=9$ ordinal features on a $1$--$10$ scale (clump thickness, uniformity of cell size and shape, marginal adhesion, etc.). We drop the sample-code-number ID column, remove the $16$ rows that contain native ``\texttt{?}'' values, and use the resulting $n=683$ samples. 

\textbf{Gisette} is a high-dimensional binary classification benchmark constructed from handwritten digit images. We stack the official train, validation, and unlabeled test splits to obtain $n=13{,}500$ samples with $p=5{,}000$ integer-valued pixel/aggregate features. Gisette serves as an ideal stress test for whether imputers remain accurate and scalable when the feature dimension is large.

\paragraph{Preprocessing.}
Before applying any imputation method, the datasets are normalized feature-wise using \texttt{StandardScaler} from \texttt{scikit-learn}. All RMSE numbers reported in the main paper are therefore on the standardized scale, which is comparable across features and across datasets.

\paragraph{MNAR mask generation.}
Real-world datasets do not come with ground-truth missing entries that allow us to evaluate imputation accuracy. We therefore inject controlled MNAR missingness on top of the full data matrix, so that the original observed values can serve as the ground truth for the masked entries. Specifically, for each dataset we form a self-masked logistic mechanism that depends on both each feature's own value and on a nonlinear summary of the other features. For each row $i$ and feature $j$, the score is
\begin{equation}
    s_{ij} = 0.6\,\widetilde{x}_{ij} + 0.4\,\tanh\!\left(\widetilde{\mathbf{x}}_i^{\top} W\right)_j,
    \qquad
    W \in \mathbb{R}^{p\times p},\;\; W_{kj}\sim \mathcal{N}(0,\,0.3^2/p),
\end{equation}
where $\widetilde{\mathbf{x}}_i$ is the standardized row. The observation probability is $\mathbb{P}(r_{ij}=1\mid \mathbf{x}_i) = \mathrm{sig}(s_{ij}+\beta_0)$, with the intercept $\beta_0$ calibrated by bisection so that the average per-entry observation rate matches the target (e.g., $80\%$ for missing rate $0.2$). To avoid degenerate corner cases we additionally enforce that every row has at least one observed entry and every column has at least one observed entry. The resulting mask is non-ignorable (MNAR) because larger target coordinates, and combinations of values that activate the nonlinear coupling, are systematically more likely to be missing, so any imputer that conditions only on the observed entries (and ignores the mask) is likely to be biased.

\section{Baseline methods}\label{app:baselines}

We compare MissBGM against seven imputation baselines that span classical statistical methods, optimal-transport-based methods, ensemble multiple imputation, and modern deep generative models. To ensure a fair comparison, every baseline operates on exactly the same standardized incomplete matrix described in Appendix~\ref{app:datasets-real}, sees exactly the same missingness mask as MissBGM, and uses the same random seed for shared randomness from the same data realization. All baselines are evaluated on the standardized scale, so that RMSE numbers are directly comparable across methods.

\paragraph{Mean.}
Mean imputation replaces each missing entry by the mean of the corresponding feature computed over its observed entries. It is parameter-free, computationally efficient, and serves as the simplest baseline for imputation. We use a \texttt{numpy}-based implementation that operates directly on the standardized matrix.

\paragraph{Optimal Transport (OT)}
OT-based imputation casts missing data completion as the problem of minimizing the entropy-regularized Sinkhorn divergence between random mini-batches of the (partially imputed) data, so that the imputed values produce a self-consistent empirical distribution. We use the \texttt{OTimputer} implementation from the ForestDiffusion benchmark codebase. The regularization parameter $\varepsilon$ is selected automatically using the authors' \texttt{pick\_epsilon} heuristic with quantile $0.5$ and scale factor $0.05$. Each imputation runs the Sinkhorn loop for $3000$ iterations with batch size $128$ and Adam learning rate $10^{-2}$. To produce $K$ multiple imputations ($K=5$ as default unless otherwise specified), we re-run the procedure $K$ times with independent random seeds and aggregate the results by averaging as the point estimate.

\paragraph{Iterative Conditional Equations (ICE)}
ICE (also known as MICE-PMM in its multiple-imputation form) repeatedly regresses each partially observed feature on all other features in a round-robin fashion, drawing imputed values from the posterior predictive distribution of the regressor. We use \texttt{IterativeImputer} from \texttt{scikit-learn} library, a maximum of $10$ outer round-robin iterations, and Bayesian Ridge regressors as the per-feature conditional models. Each of the $K$ imputations uses an independent random seed; the imputed entries are clipped to the observed marginal range to suppress out-of-range posterior draws on heavy-tailed features.

\paragraph{Miceforest}
Miceforest implements multiple imputation by chained equations using LightGBM gradient-boosted trees as the per-feature conditional models, which makes it well suited to nonlinear feature dependencies in tabular data. We use the official \texttt{miceforest.ImputationKernel} as called from the ForestDiffusion driver: \texttt{datasets=K} parallel imputation chains seeded by a shared \texttt{random\_state}. The $K$ completed datasets are then aggregated by averaging as the point estimate. $K=5$ in point estimate experiments and $K=1000$ for uncertainty quantification on the synthetic benchmark.

\paragraph{GAIN}
GAIN formulates imputation as a conditional generative adversarial learning problem: a generator network completes the missing entries while a discriminator, given a hint mask, tries to identify which entries were imputed. We use the original GAIN implementation with the default hyperparameters: batch size $128$, hint rate $0.9$, reconstruction-loss weight $\alpha=100$, and $10{,}000$ training iterations. The trained generator is then sampled $K$ times based on different latent variables to produce $K$ multiple imputations. $K=5$ in point estimate experiments and $K=1000$ for uncertainty quantification on the synthetic benchmark.

\paragraph{ForestDiffusion}
ForestDiffusion adapts score-based diffusion to tabular data by replacing the score neural network with a tree-based ensemble that learns to predict the noise injected at each diffusion step. Imputation is performed by running the reverse-time diffusion from Gaussian noise while clamping the observed entries to their true values at every denoising step. We use the default tabular configuration: an XGBoost backbone (\texttt{forest\_model=xgboost}) under the variance-preserving SDE (\texttt{diffusion\_type=vp}), with $n_t=50$ diffusion time steps for both training and sampling, $\beta_{\min}=0.1$, $\beta_{\max}=8.0$, $\varepsilon=10^{-3}$, max tree depth $7$, $100$ trees per step, learning rate $0.3$, and a duplication factor of $100$ for the noise/time pair augmentation that ForestDiffusion uses to stabilize training. Sampling produces $K$ completed matrices that are then averaged. $K=5$ in point estimate experiments and $K=1000$ for uncertainty quantification on the synthetic benchmark.

\paragraph{TabCSDI}
TabCSDI is a conditional score-based diffusion model adapted to tabular data, in which the score network is a transformer trained to denoise the missing entries conditioned on the observed ones. Imputation is performed by running the conditional reverse diffusion to draw posterior samples of the missing entries. We use the official TabCSDI implementation, retaining the default network architecture, training schedule, and noise schedule, and configure the model with one numerical-feature group covering all standardized columns. $K$ posterior completions are sampled per test matrix and averaged. $K=5$ in point estimate experiments and $K=1000$ for uncertainty quantification on the synthetic benchmark.

\section{Experimental details}\label{app:implementation}

This appendix section provides additional experimental details regarding model training, testing, hyperparameters, and environment. To ensure a fair comparison across methods, we control the randomness in each dataset so that MissBGM and all baselines are trained and evaluated on exactly the same observed entries and missingness mask under the same random seed.

\subsection{Model architecture}

\paragraph{Data-generating network $g_\theta$.}
The conditional distribution $p_\theta(\mathbf{x}\mid\mathbf{z})$ is parameterized by a fully connected network (FCN) that maps $\mathbf{z}\in\mathbb{R}^{d}$ to a diagonal Gaussian over $\mathbf{x}\in\mathbb{R}^{p}$: it outputs a mean vector $\boldsymbol{\mu}_\theta(\mathbf{z})$ and a positive variance vector $\boldsymbol{\sigma}^2_\theta(\mathbf{z})$ (softplus on the variance head plus a small floor). Hidden layers use LeakyReLU ($\alpha=0.2$) after each dense layer, batch normalization on the input to the first hidden block, and L2 weight decay ($10^{-4}$) on kernels and biases. $g_\theta$ has five layers and each layer has $120$ hidden units as default.

\paragraph{Missingness network $f_\phi$.}
The Bernoulli logits for $\mathbf{r}\mid\mathbf{x}$ are produced by a separate FCN with the same LeakyReLU/L2 structure as above. $f_\phi$ has two layers and each layer has $64$ hidden units as default.

\paragraph{Bayesian neural network for $g_\theta$ and $f_\phi$.}
We provide an optional Bayesian neural network (BNN) implementation to explicitly account for model-parameter uncertainty for both $g_\theta$ and $f_\phi$. In this variant, deterministic dense layers are replaced by \texttt{DenseFlipout} layers from TensorFlow Probability. \texttt{DenseFlipout} performs variational Bayes with weight posteriors by sampling pseudo-independent sign-perturbations per example (Flipout), which yields an unbiased stochastic gradient estimator with substantially reduced gradient variance compared with naive shared-weight reparameterization. In practice, this variance reduction makes optimization more stable on mini-batches, improves training signal-to-noise, and provides better calibrated uncertainty in predictions when posterior weight samples are propagated through the imputation pipeline.

\subsection{EGM initialization}

Before the alternating stochastic optimization, we use an encoding generative modeling (EGM) stage~\citep{liu2024encoding} to obtain a stable warm start for both network parameters and latent representations. The key idea is to additionally introduce an encoder network for pretraining a coupled encoder-generator pair so that latent samples and reconstructed data are already approximately aligned in distribution, instead of starting from random weights and random latent spaces. Specifically, we first apply a simple KNN imputer with five nearest neighbors to fill the missing entries, then an auxiliary encoder network $e(\mathbf{x})$ is introduced to map from the complete data space to the latent space, and a generator network $g(\mathbf{z})$ is introduced to map the latent samples back to the data space $\mathbf{x}$. The encoder and generator are trained jointly to minimize the adversarial loss and the cycle-consistency loss. Through adversarial matching, we desire that $e(\mathbf{x})$ should match the prior distribution of the latent variables $\pi(\mathbf{z})$, which is a standard normal distribution. The cycle-consistency loss encourages that $g(e(\mathbf{x}))$ should be close to $\mathbf{x}$.

Conceptually, this EGM stage encourages the encoder and generator to learn a coherent inverse relationship, where the encoder can be seen as a shortcut function as opposed to Bayesian inference of latent variables. The EGM stage is only used to initialize the model parameters and latent variables, and the encoder $e(\mathbf{x})$ will be discarded during the alternating stochastic optimization.

We only run EGM initialization for $1{,}500$ mini-batches as default, which can be typically done within $1$ minute. Then we initialize each sample-specific latent variable with a forward encoder pass,
\begin{equation}
    \mathbf{Z}^{(0)} = e(\mathbf{X}),
\end{equation}
and use these latents together with the pretrained generator as the starting point of Algorithm~\ref{alg:missbgm-train}. Empirically, this warm start improves optimization stability, speeds up convergence of the alternating updates for $(\mathbf{Z},\mathbf{X}_{\mis},\theta,\phi)$, and helps avoid poor local solutions that occur more often with purely random initialization.

\subsection{Alternating stochastic optimization}

MissBGM is trained by blockwise stochastic optimization on mini-batches. The procedure alternates between sample-specific updates, which refine the latent variables and the missing entries for the current batch, and global updates, which refine the parameters of the data model and the missingness model. This alternating structure mirrors the factorization of the joint posterior and avoids the need to optimize all unknowns simultaneously. More importantly, the alternating stochastic optimization only requires a single sample or a mini-batch of samples in each updating step, which makes it computationally efficient and scalable for large datasets.

We begin by running the EGM warm-start stage described above, which provides the initial state for the alternating updates. For a given batch during training, we first perform several local refinement steps on the latent variables and missing entries while holding the global network parameters fixed. The latent update maximizes the conditional log-posterior likelihood of the latent variables, while the missing-data update maximizes the conditional log-posterior likelihood of the missing entries under the joint data and mask model. After each missing-data update, the observed coordinates are projected back to their original values so that only the unobserved entries are allowed to move.

After the local refinement stage, we update the global parameters of $g_\theta$ and $f_\phi$ using the current completed batch. All gradient-based updates of the latent variables, the missing entries, and the global model parameters are performed with the Adam optimizer~\citep{kingma2014adam}. All batch losses are normalized by the feature dimension for numerical stability, and gradient clipping is used in the sample-specific steps to stabilize optimization. When Bayesian neural networks are used, the objective additionally includes the KL regularization terms associated with the variational weight posterior. Otherwise the updates reduce to stochastic gradient descent. In all cases, training proceeds by repeating this local-global alternation until the number of epochs is reached. The detailed algorithm is summarized in Algorithm~\ref{alg:missbgm-train}.

\begin{algorithm}[t]
\caption{Alternating stochastic optimization for MissBGM.}
\label{alg:missbgm-train}
\begin{algorithmic}[1]
\Require Incomplete data matrix $\mathbf{X}_{\mathrm{obs}}$, observation mask $\mathbf{R}$, number of epochs $T$, mini-batch size $B$, number of inner refinement steps $K$, learning rates $\eta_1$ for latent-variable and missing-value updates and $\eta_2$ for global parameter updates 
\State Run EGM warm-start to initialize the model parameters and $(\mathbf{Z}^{(0)},\mathbf{X}^{(0)})$
\State Set the initial state $(\mathbf{Z},\mathbf{X}) \gets (\mathbf{Z}^{(0)},\mathbf{X}^{(0)})$
\For{$t=1,2,\ldots,T$}
    \State Randomly shuffle the data and split them into mini-batches $\mathcal{B}$ of size at most $B$
    \For{each mini-batch $\mathcal{B}$}
        \For{$k=1,2,\ldots,K$}
            \State Update the batch latent variables by
            \[
                \mathbf{Z}_{\mathcal{B}} \leftarrow \mathbf{Z}_{\mathcal{B}} - \eta_1\, \nabla_{\mathbf{Z}_{\mathcal{B}}} \mathcal{L}_{z}(\mathcal{B}),
            \]
            \Statex\hspace{1.5em}where $\mathcal{L}_{z}(\mathcal{B})$ is the mini-batch latent objective induced by Eq.~\eqref{eq:z_objective}
            \State Update the batch missing entries by
            \[
                \mathbf{X}_{\mis,\mathcal{B}} \leftarrow \mathbf{X}_{\mis,\mathcal{B}} - \eta_1\, \nabla_{\mathbf{X}_{\mis,\mathcal{B}}} \mathcal{L}_{x_{\mis}}(\mathcal{B}),
            \]
            \Statex\hspace{1.5em}where $\mathcal{L}_{x_{\mis}}(\mathcal{B})$ is the mini-batch missing-data objective induced by Eq.~\eqref{eq:xmis_objective}
            
        \EndFor
        \State Update the data-model parameters by stochastic gradient ascent on the ELBO,
        \[
            \theta \leftarrow \theta + \eta_2\, \nabla_{\theta} \mathcal{L}_{\theta}(\mathcal{B}),
        \]

        \State Update the missingness-model parameters by
        \[
            \phi \leftarrow \phi + \eta_2\, \nabla_{\phi} \mathcal{L}_{\phi}(\mathcal{B}),
        \]
        \Statex\hspace{1.5em}where $\mathcal{L}_{\theta}$ and $\mathcal{L}_{\phi}$ are the corresponding mini-batch ELBO in Eq.~\eqref{eq:elbo_bnn}

    \EndFor
\EndFor
\State \Return fitted parameters $(\theta,\phi)$ and final state $(\mathbf{Z},\mathbf{X})$
\end{algorithmic}
\end{algorithm}

\subsection{Posterior inference and uncertainty quantification}

\paragraph{Inference on the training data.}
In our imputation setting, posterior inference is most often performed on the same incomplete dataset used for training. In this case, no additional optimization is required. The alternating stochastic optimization already returns a completed data matrix at convergence, and this terminal state serves as the default point estimate. If posterior uncertainty is further required, we directly launch MCMC from the fitted terminal state while keeping the global parameters fixed.

\paragraph{Inference on a new dataset.}
When the test data differ from the training data, we keep the fitted global parameters $(\theta,\phi)$ unchanged and update only the sample-specific latent variables and missing entries for the new dataset. Starting from a simple initial completion, we run the same local alternating refinement steps as in training, namely latent updates and missing-data updates, but without modifying the global model parameters. This produces a dataset-specific MAP state, which serves both as the point estimate for the new incomplete data and as the starting point for posterior sampling when uncertainty quantification is requested.

\paragraph{Posterior sampling.}
Given either the fitted training state or the dataset-specific MAP state, we approximate the conditional posterior by an HMC-within-Gibbs sampler. Each sweep alternates between updating the latent variables conditional on the current completed data and updating the missing entries conditional on the current latent state and observed coordinates. The resulting sampler is therefore a posterior refinement of the deterministic estimate produced by alternating stochastic optimization, rather than a separate inference procedure.

\paragraph{Posterior summaries.}
After discarding burn-in iterations, we retain Monte Carlo draws of the completed data matrix and summarize them through posterior means and empirical prediction intervals for the missing entries. In the main experiments, we use $1000$ burn-in iterations and retain $1000$ posterior draws, with initial step size $0.1$ and $5$ leapfrog steps.

\subsection{Model hyperparameters}

Table~\ref{tab:missbgm-hparams} summarizes the main MissBGM hyperparameters used in the main experiments. Dataset-specific YAML files may override few entries for particular small or large datesets. For example, we use $\texttt{z\_dim=10}$, $\texttt{egm\_init.n\_iter=5000}$ for the largest Gisette dataset, and use $\texttt{g\_units=[120,120,120]}$ for the smallest Wine dataset to keep the model simple and stable. The \texttt{epochs} is set to $50$ for all simulation benchmark experiments. Overall, the model is robust to the choice of hyperparameters for datasets with varying sample sizes and feature dimensions.

The EGM encoder architecture is chosen as a rough mirror of the generator \(g_\theta\) (e.g., reversing the layer widths in \texttt{g\_units}), which stabilizes initialization and reduces the need for manual tuning. When the underlying data distribution is complex (large \(p\), strong nonlinearity, or multi-modality), using a larger latent dimension \texttt{z\_dim} makes more sense to improve fit and downstream imputation quality, at the cost of additional compute. In our experiments, we find that setting $z_{\text{dim}}=5$ by default is across-the-board effective for most datasets.

\begin{table}[t]
    \centering
    \small
    \caption{MissBGM main hyperparameters.}
    \label{tab:missbgm-hparams}
    \begin{tabular}{@{}p{0.23\textwidth}p{0.49\textwidth}p{0.16\textwidth}@{}}
        \toprule
        \textbf{Hyperparameter} & \textbf{Explanation} & \textbf{Default} \\
        \midrule
        \texttt{z\_dim} & Dimension of the latent variable $\mathbf{z}$. & $5$ \\
        \texttt{epochs} & Number of outer training epochs. & $200$ \\
        \texttt{batch\_size} & Minibatch size used for stochastic training updates. & $16$ \\
        \texttt{beta} & Weight on the missingness-model term in Eq.~\eqref{eq:xmis_objective}. & $0.01$ \\
        \texttt{g\_units} & Hidden-layer widths of $g_\theta$. & $[120,120,120,120,120]$ \\
        \texttt{missingness\_units} & Hidden-layer widths of $f_\phi$. & $[64,64]$ \\
        \texttt{lr\_theta} & Learning rate for updating $\theta$. & $0.005$ \\
        \texttt{lr\_phi} & Learning rate for updating $\phi$. & $0.005$ \\
        \texttt{lr\_z} & Learning rate for updating $\mathbf{Z}$. & $0.002$ \\
        \texttt{lr\_x} & Learning rate for updating $\mathbf{X}_{\mis}$. & $0.002$ \\
        \texttt{n\_inner\_steps} & Number of inner updates of $\mathbf{Z}$ and $\mathbf{X}_{\mis}$ per minibatch. & $3$ \\
        \texttt{use\_bnn} & Whether to use BNNs & \texttt{false} \\
        \texttt{kl\_weight} & Weight $\tau$ on the KL regularization & $5\times 10^{-5}$ \\
        \texttt{test\_epochs} & Number of test epochs before MCMC refinement. & $30$ \\
        \texttt{egm\_init.enabled} & Whether to apply EGM initialization. & \texttt{true} \\
        \texttt{egm\_init.n\_iter} & Number of mini-batches in EGM initialization. & $1500$ \\
        \texttt{egm\_init.e\_units} & Hidden-layer widths of encoder in EGM. & $[120,120,120,120,120]$ \\
        \texttt{egm\_init.dz\_units} & Hidden-layer widths of latent discriminator in EGM. & $[64,32,8]$ \\
        \texttt{egm\_init.dx\_units} & Hidden-layer widths of data discriminator in EGM. & $[64,32,8]$ \\
        \texttt{posterior.alpha} & Significance level for posterior prediction intervals. & $0.05$ \\
        \texttt{posterior.n\_mcmc} & Number of retained posterior draws after burn-in. & $1000$ \\
        \texttt{posterior.burn\_in} & Number of warm-up iterations in MCMC. & $1000$ \\
        \texttt{posterior.step\_size} & Initial step size in MCMC. & $0.1$ \\
        \bottomrule
    \end{tabular}
\end{table}

\subsection{Experimental environment}

MissBGM was implemented in \texttt{Python} using \texttt{TensorFlow} for neural network modeling and alternating stochastic optimization, and \texttt{TensorFlow Probability} (TFP)~\citep{dillon2017tensorflow} for Bayesian neural networks and HMC-within-Gibbs posterior sampling. Note that the posterior inference for one sample is independent of other samples, so the Markov chains of all test samples are run in parallel using TFP for efficiency. All benchmark experiments were conducted on a Linux-based high-performance computing cluster managed by \texttt{Slurm}. The GPU compute nodes used in our experiments were equipped with NVIDIA RTX 5000 Ada GPUs and dual-socket CPUs (48 CPU cores per node in total). For each experiment, we requested a maximum wall-clock time of 7 days and 300\,GB of memory. We only used one GPU per experiment for deep learning models.

\section{Software usage for MissBGM}\label{app:software-usage}
This section provides an example showing how to use MissBGM on a synthetic dataset. The code first loads a YAML configuration file into \texttt{params}, which contains model and inference hyperparameters such as network widths, learning rates, and MCMC settings, etc. It then prepares an incomplete dataset \texttt{data}: the main input to MissBGM is the partially observed matrix \texttt{x\_obs}\(\in\mathbb{R}^{n\times p}\), with missing entries encoded as \texttt{np.nan}; the optional binary mask \texttt{mask}\(\in\{0,1\}^{n\times p}\) records which entries are observed (will be inferred from \texttt{x\_obs} if not provided), and the optional complete data matrix \texttt{x\_full} for evaluation only. The call \texttt{MissBGM(params, random\_seed=42)} initializes the model, and \texttt{model.fit(...)} trains it using \texttt{x\_obs} and \texttt{mask}. After fitting, \texttt{model.x\_map\_imputed\_} stores the MAP completed data matrix, while \texttt{model.predict(...)} returns posterior summaries, including posterior-mean imputations and prediction intervals for all the missing entries. Note that the argument \texttt{x\_true} is only used for internal evaluation and can be set to \texttt{None}. The argument \texttt{mask} can be automatically inferred from \texttt{x\_obs} by setting it to \texttt{None}.

The example below illustrates running MissBGM on the synthetic benchmark, including model instantialization, model training and posterior-sampling inference.

\begin{codeblock}{Example: run MissBGM on a synthetic dataset}
import yaml
from missbgm.models import MissBGM
from missbgm.datasets import simulate_mnar_oracle_data

params = yaml.safe_load(open("configs/MNAR_oracle.yaml", "r"))
data = simulate_mnar_oracle_data(
    n_samples=500,
    x_dim=50
)

# Instantiate the MissBGM model with the configuration
model = MissBGM(params, random_seed=42)

# Train the MissBGM model
model.fit(data=data["x_obs"], mask=data["mask"], x_true=data["x_full"])

# Get the MAP imputation
map_imputed = model.x_map_imputed_

# Make posterior predictions with uncertainty quantification
mcmc_imputed, intervals = model.predict(
    data=data["x_obs"],
    mask=data["mask"],
    x_true=data["x_full"],
    alpha=0.05
)
\end{codeblock}

In this example, \texttt{simulate\_mnar\_oracle\_data} returns the observed matrix \texttt{x\_obs} (the missing entries are all \texttt{np.nan}), the binary missingness mask \texttt{mask}, and the complete synthetic ground truth \texttt{x\_full}. The call to \texttt{model.fit(...)} runs EGM initialization, followed by the alternating stochastic optimization procedure. The subsequent call to \texttt{model.predict(...)} returns the posterior mean imputation and the corresponding posterior prediction intervals for the missing entries. Note that the argument \texttt{x\_true} is only used for internal evaluation and can be set to \texttt{None}. The argument \texttt{mask} can be automatically inferred from \texttt{x\_obs} by setting it to \texttt{None}.

When posterior inference is performed on the same incomplete dataset used for training, as in the example above, MissBGM directly reuses the fitted training state and does not re-optimize the global parameters. If uncertainty quantification is not needed, one may stop after training and use the fitted state as the default point estimate. If uncertainty quantification is desired, posterior sampling can then be launched through \texttt{predict(...)}.

%\section{Additional experiments}\label{app:additional-experiments}

\section{Effect of missing rate and sample size}\label{app:missing-rate-size}

To complement Figure~\ref{fig:robustness}(b), which reports the missing-rate experiment at $n=5000$, we provide the corresponding curves for three additional sample sizes, namely $n\in\{500,1000,10000\}$. In all cases, we fix $p=50$ and vary the missing rate only. These additional plots show that the qualitative conclusions are stable across sample sizes. MissBGM remains the strongest method overall.

A second consistent pattern is that the RMSE curves are not strictly monotone in the missing rate. Instead, most methods exhibit a broad U-shaped trend, with the lowest RMSE often attained at moderate missingness levels and larger errors reappearing at high missingness rates. This behavior is consistent with the synthetic data construction and the discussion in Appendix~\ref{app:datasets-synthetic}. Overall, the figures below reinforce the conclusion from Figure~\ref{fig:robustness}(b): MissBGM is robust to substantial changes in missingness severity.

\begin{figure}[t]
    \centering
    \includegraphics[width=0.82\linewidth]{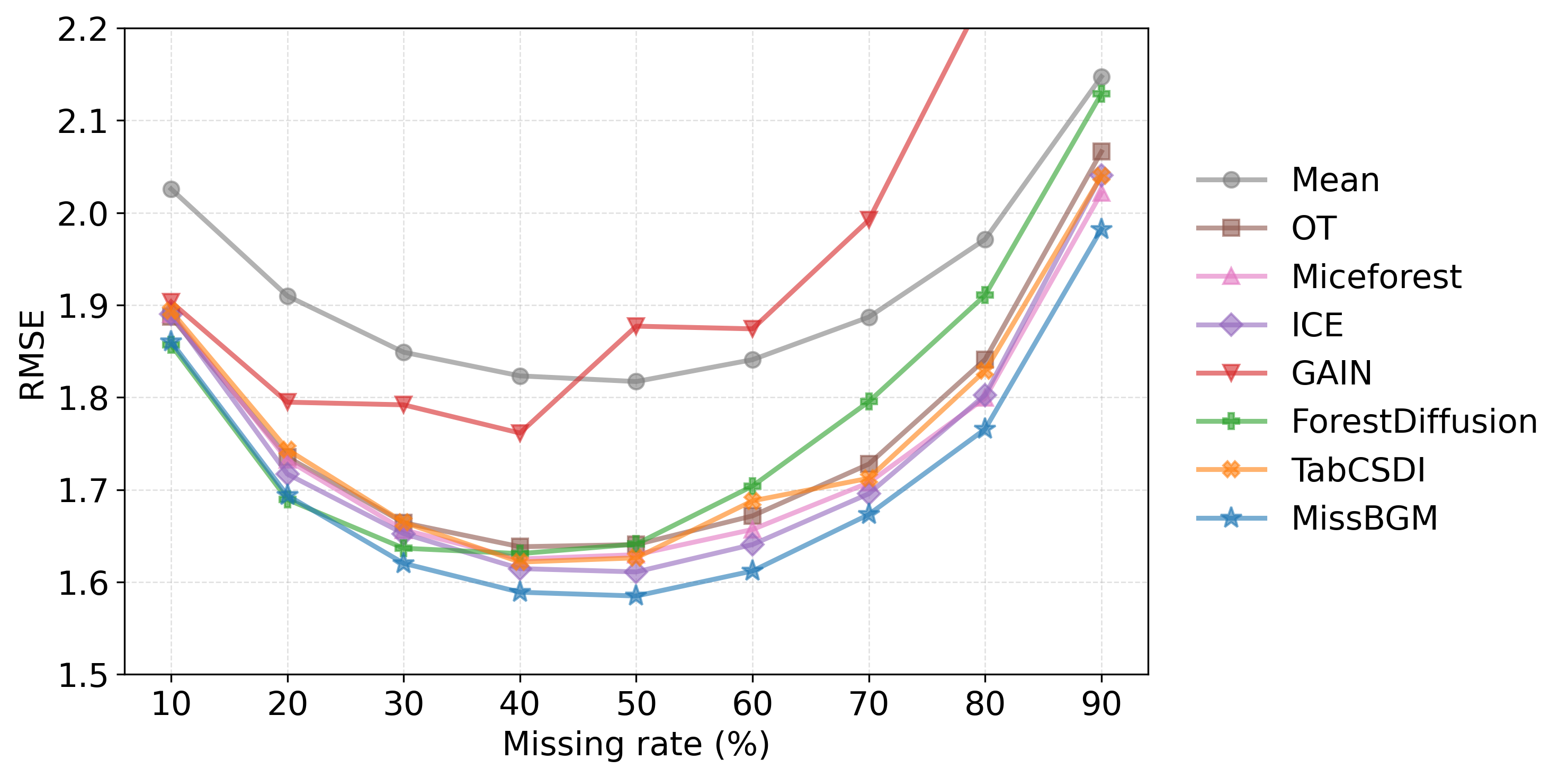}
    \caption{RMSE with different missing rate on the synthetic benchmark with $n=500$ and $p=50$.}
    \label{fig:missing-rate-500}
\end{figure}

\begin{figure}[t]
    \centering
    \includegraphics[width=0.82\linewidth]{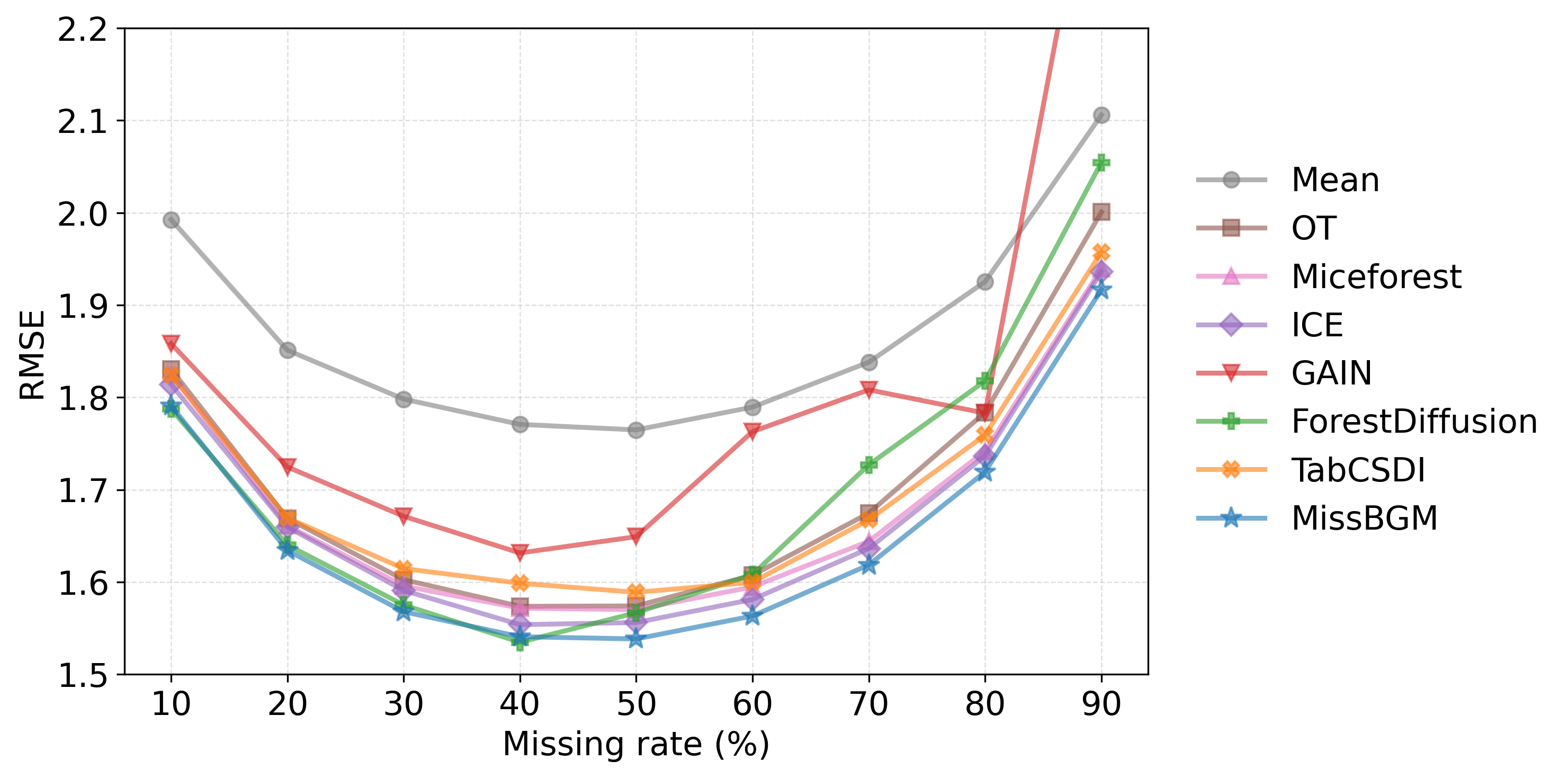}
    \caption{RMSE with different missing rate on the synthetic benchmark with $n=1000$ and $p=50$.}
    \label{fig:missing-rate-1000}
\end{figure}

\begin{figure}[t]
    \centering
    \includegraphics[width=0.82\linewidth]{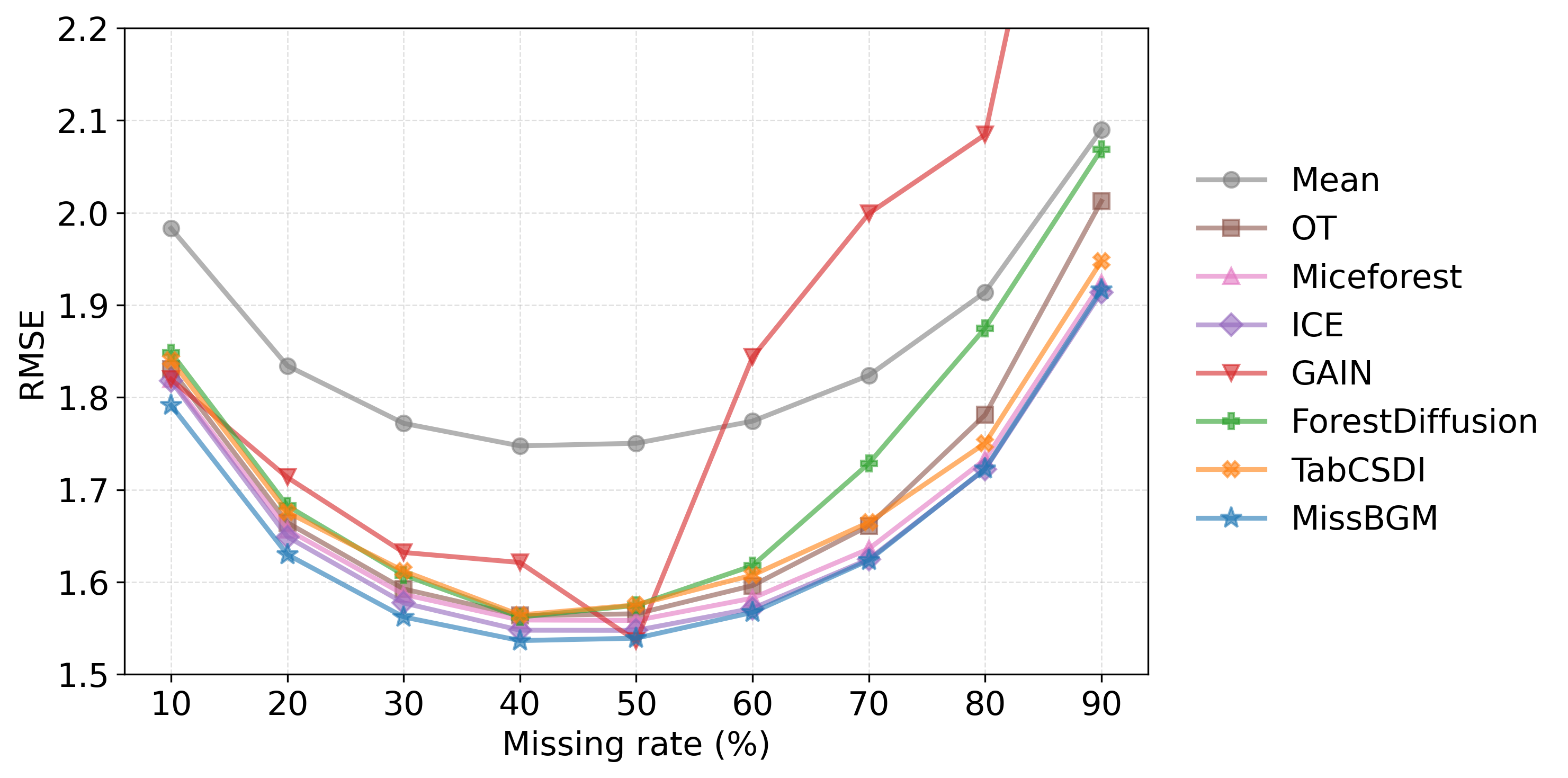}
    \caption{RMSE with different missing rate on the synthetic benchmark with $n=10000$ and $p=50$.}
    \label{fig:missing-rate-10000}
\end{figure}

\section{Running time}\label{app:running-time}

\begin{figure}[t]
    \centering
    \includegraphics[width=0.80\linewidth]{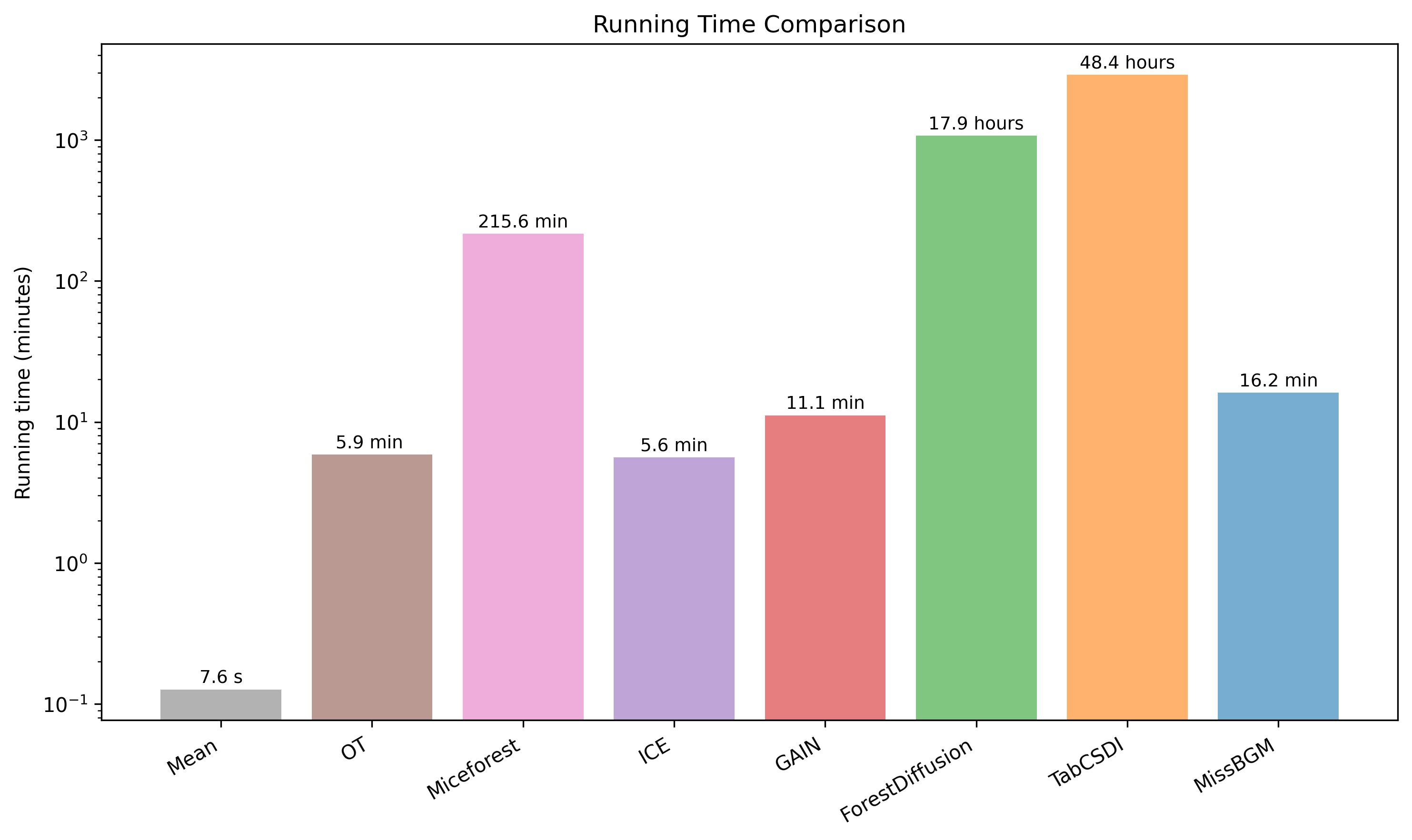}
    \caption{Running time comparison across imputation methods on a representative synthetic benchmark with moderate complexity settings ($n=5000$, $p=200$).}
    \label{fig:running-time}
\end{figure}

Figure~\ref{fig:running-time} reports the running time of all compared methods on a representative synthetic benchmark with moderate complexity settings ($n=5000$, $p=200$, and missing rate $r=0.2$). The y-axis is shown on a logarithmic scale as running time spans different orders of magnitude, from seconds to days. Mean imputation finishes in seconds while OT and ICE both complete in less than 10 minutes. Miceforest requires $215.6$ minutes, which is already more than one order of magnitude slower than MissBGM due to the repeated fitting of conditional models for each feature across multiple iterations. The diffusion-based methods are substantially more expensive. ForestDiffusion requires $17.9$ hours and TabCSDI requires $48.4$ hours. 

Therefore, although MissBGM jointly models both the data-generating process and the missingness mechanism and additionally supports uncertainty quantification, its computational cost remains moderate relative to other deep learning-based imputers. In particular, MissBGM is roughly comparable to GAIN, but substantially faster than miceforest and diffusion-based methods.

\section{Effect of Bayesian neural networks}\label{app:bnn}

To complement the main experiments, we evaluate MissBGM with Bayesian neural networks (BNNs) on the synthetic benchmark. We use the same data-generating process and experimental protocol as in Table~\ref{tab:synthetic-rmse}, but replace the deterministic networks $g_{\theta}$ and $f_{\phi}$ with their BNN counterparts (implemented via variational Bayes) by setting \texttt{use\_bnn=True}. We report results for three missing rates $r=0.3$ and sample sizes $n\in\{500,1000,5000,10000\}$. Overall, the BNNs achieve comparable or slightly worse performance than the deterministic networks.

\begin{table}[t]
    \centering
    \caption{Effect of Bayesian neural networks (BNNs) on the synthetic benchmark at missing rate $r=0.3$. We report mean (s.d.) RMSE over 10 runs.}
    \label{tab:synthetic-bnn-03}
    \setlength{\tabcolsep}{7pt}
    \renewcommand{\arraystretch}{1.05}
    \small
    \begin{tabular}{@{}lcccc@{}}
        \toprule
        \textbf{Method} & \textbf{Size=500} & \textbf{Size=1000} & \textbf{Size=5000} & \textbf{Size=10000} \\
        \midrule
        MissBGM w/o BNN & \textbf{1.620 (0.011)} & \textbf{1.568 (0.009)} & 1.599 (0.010) & \textbf{1.562 (0.006)} \\
        MissBGM w/ BNN  & 1.632 (0.017)     & 1.570 (0.012)     & \textbf{1.592 (0.012)}     & 1.564 (0.008) \\
        \bottomrule
    \end{tabular}
\end{table}

\newpage

\end{document}